\documentclass[aoap]{imsart}

\RequirePackage{amsthm,amsmath,amsfonts,amssymb}
\RequirePackage[numbers]{natbib}
\usepackage{xr}
\usepackage{amsmath,amssymb,mathrsfs,amsthm,amsfonts}
\usepackage[inline]{enumitem} 
\usepackage[usenames,dvipsnames]{xcolor}
\usepackage{hyperref}
\usepackage{graphicx}
\usepackage{pgf}
\usepackage{tikz}
\usepackage{tikz-cd}
\usetikzlibrary{arrows, matrix, arrows.meta, cd}
\usepackage[textsize=tiny]{todonotes}
\usepackage{comment}
\hypersetup{%
	colorlinks=true, linkcolor=blue,
	citecolor=ForestGreen
}
\usepackage{algorithm}      
\usepackage{algpseudocode}
\usepackage[noabbrev,capitalize,nameinlink]{cleveref}
\startlocaldefs
\numberwithin{equation}{section}
\newtheorem{theorem}{Theorem}[section]
\newtheorem{proposition}[theorem]{Proposition}
\newtheorem{lemma}[theorem]{Lemma}
\newtheorem{claim}[theorem]{Claim}

\newtheorem*{question*}{Question}

\theoremstyle{definition}
\newtheorem{definition}[theorem]{Definition}

\newtheorem{question}[theorem]{Question}
\newtheorem*{definition*}{Definition}

\theoremstyle{remark}

\allowdisplaybreaks
\usepackage{acronym}
\usepackage{tikz}
\usetikzlibrary{matrix,graphs,arrows,positioning, shapes, calc,decorations.markings,decorations.pathmorphing,shapes.symbols,hobby}

\acrodef{KPZ}{Kardar--Parisi--Zhang}
\acrodef{SHE}{Stochastic Heat Equation}
\acrodef{LDP}{Large Deviation Principle}

\newcommand{\N}{\mathbb{N}}
\newcommand{\R}{\mathbb{R}} 
\newcommand{\Z}{\mathbb{Z}} 

\renewcommand{\hat}{\widehat}

\usepackage{graphicx}

\numberwithin{equation}{section}

\setenumerate[1]{label=(\arabic*)}
\makeatletter
\newcommand{\biggg}{\bBigg@{4}}
\newcommand{\Biggg}{\bBigg@{5}}
\makeatother

\renewcommand{\S}{\mathbb{S}}

\newcommand{\1}{\mathbf{1}}
\renewcommand{\P}{\mathbb{P}}

\newcommand{\E}{\mathbb{E}}
\newcommand{\B}{\mathbb{B}}
\renewcommand{\S}{\mathbb{S}}

\newcommand{\iid}{\stackrel{iid}{\sim}}

\usepackage{comment}
\DeclareMathOperator{\Unif}{Unif}

\let \Re \relax
\DeclareMathOperator{\Re}{ReLU}
\let \N \relax
\DeclareMathOperator{\N}{N}

\DeclareMathOperator{\dist}{dist}

\DeclareMathOperator{\conv}{conv}

\numberwithin{equation}{section}

\endlocaldefs

\begin{document}

\begin{frontmatter}
\title{Randomly Initialized One-Layer Neural Networks Make Data Linearly Separable}
\runtitle{One-Layer RINNs Make Data Linearly Separable}

\begin{aug}
\author[A]{\fnms{Promit} \snm{Ghosal}\ead[label=e1]{promit@brandeis.edu}}\and
\author[B]{\fnms{Srinath} \snm{Mahankali}\ead[label=e2]{srinathm@mit.edu}}\and
\author[C]{\fnms{Yihang} \snm{Sun}\ead[label=e3]{kimisun@mit.edu}}
\address[A]{Department of Mathematics, Brandeis University, \printead{e1}
	}

\address[B]{Department of Mathematics, MIT, \printead{e2}
	}
 \address[C]{Department of Mathematics, MIT, \printead{e3}
	}
\end{aug}

\begin{abstract}
Recently, neural networks have demonstrated remarkable capabilities in mapping two arbitrary sets to two linearly separable sets. The prospect of achieving this with randomly initialized neural networks is particularly appealing due to the computational efficiency compared to fully trained networks. This paper contributes by establishing that, given sufficient width, a randomly initialized one-layer neural network can, with high probability, transform two sets into two linearly separable sets without any training. Moreover, we furnish precise bounds on the necessary width of the neural network for this phenomenon to occur. Our initial bound exhibits exponential dependence on the input dimension while maintaining polynomial dependence on all other parameters. In contrast, our second bound is independent of input dimension, effectively surmounting the curse of dimensionality. The main tools used in our proof heavily relies on a fusion of geometric principles and concentration of random matrices.

\end{abstract}



\end{frontmatter}

\tableofcontents


\section{Introduction}

Over the past decade, neural networks (NNs) have enjoyed immense success across a multitude of applications, as evidenced by seminal works such as \cite{lecun2015deep, schmidhuber2015deep, goodfellow2016deep}. Notably, the practice of random initialization has substantially bolstered the capabilities of NNs, rendering computations more cost-effective than employing fully trained NNs, as demonstrated by \cite{he2015delving, goodfellow2016deep, arpit2019benefits}.

Within the realm of data classification, the preliminary step of employing a randomly initialized neural network (RINN) in conjunction with a standard linear classifier has exhibited remarkable performance \cite{rahimi2008weighted, zhang2016understanding}. However, a rigorous theoretical underpinning for this phenomenon has remained largely uncharted until very recently, as alluded to in \cite{an2015can, dirksen2021separation}.

This paper delves into the realm of linear separability in datasets through the lens of RINNs, seeking to provide a theoretical foundation for this intriguing and practically valuable approach. We formulate this key problem as follows.
\begin{question}
Consider two bounded, possibly infinite sets $X^{+}, X^{-}\subset \mathbb{R}^d$ that are \emph{$\delta$-separated,} i.e., $\Vert x^{+}-x^{-}\Vert \geq \delta$ for all $x^{+}\in X^{+}$ and $x^{-}\in X^{-}$. Let $F: \mathbb{R}^d \to \mathbb{R}^{n}$ denote the function defined by a feedforward neural network. The fundamental problem of data separability asks if $F(X^{+})$ and $F(X^{-})$ are \emph{linearly separable}, i.e., if there exists a nonzero vector $a\in \R^n$ and constant $c$ such that
\[{a^{\intercal}F(x)+c}=\begin{cases}
> 0 &\textup{if }x\in X^{+}\\
< 0 &\textup{if }x\in X^{-}
\end{cases},\]
in which case the \emph{margin of separation} is $\inf\lbrace \vert a^{\intercal}y+c\vert/\Vert a\Vert :y\in F(X^{+}\cup X^{-})\rbrace$.
\end{question}

We underscore that $\delta$-separation condition for $X^+$ and $X^-$ is quite permissive and holds true for most classification scenarios. The only exceptions occur in cases where a single sample could legitimately belong to both classes, making this condition applicable to a wide array of practical problems. Therefore, this condition serves as a litmus test for probing the geometric foundations of neural networks (NNs), shedding light on their separation capabilities.

In this paper, we unveil a striking result: a one-layer RINN, when endowed with sufficient width, adeptly addresses this challenge. Specifically, it effectively transforms $\delta$-separated sets into linearly separable ones with high probability. Notably, our width requirement hinges on polynomial dependency with respect to the random distribution parameters of the RINN. However, in our first principal finding, it exhibits exponential scaling concerning the input dimension. Encouragingly, our second discovery eliminates this exponential dimensionality dependence, enhancing the applicability and practicality of our approach.

\subsection{Setup and Notations}
For any $n\in \mathbb{N}$, let $[n]:=\lbrace 1, 2, \dots , n\rbrace$. Let $\Vert\cdot\Vert$ be the Euclidean norm.  Let $\B_{2}^d(c,R) = \{x\in \mathbb{R}^d:\Vert x-c\Vert \le R\}$. We use the shorthand $R\B_{2}^{d}:=\B_{2}^d(0,R)$. Let the unit sphere $\S^{d-1}$ be the surface of $\B_{2}^{d}$. Also, define the distance between any point $x\in \mathbb{R}^d$ and a set $S\subset \mathbb{R}^d$ as $\dist(x,S) :=\inf \lbrace \|x-y\|:y\in S\rbrace$.

In this paper, our standalone choice of activation function is the \emph{rectified linear activation function} ($\Re$) defined by $\Re (x) = x\1_{\lbrace x\ge 0\rbrace}$ and extended to $\R^{n}$ coordinate-wise. We denote the standard Gaussian matrix and vector where each entry is independent and identically distributed (i.i.d.) as $\N(0,1)$ by $\N(0, I_{n\times d})$ and $\N(0, I_{d})$ with appropriate dimensions. We sometimes refer to NNs by the function it defines. For a one-layer NN, it is $\Phi:\mathbb{R}^d \to \mathbb{R}^{n}$ where $\Phi (x)=\Re(Wx+b)$ for weights $W\in\R^{n\times d}$ and bias $b\in\R^n$.

For two arbitrary sets $X^{+}, X^{-} \subset R\B_{2}^{d}$, we use the notion of \emph{mutual complexity} indexed by separation parameter $\delta$ and mutual metric entropy parameter $\mu$ as in \cite[Section~7]{Ver18}. Our definition below is influenced by that in \cite{dirksen2021separation}.

\begin{definition}\label{def:complexity}
We say  two $\delta$-separated sets $X^{-}, X^{+}\subset R\B_{2}^{d}$ have \emph{$( \delta, \mu )$-mutual complexity} $N$, if $N=N^- + N^+$ is the minimum positive integer such that there exist two finite sets of \emph{centers} $C^+ = \{c_i^+\}_{i=1}^{N^+}\subset \R_2^d$ and $C^- = \{c_i^-\}_{i=1}^{N^-}\subset \R^d$, and two sets of constants $r_1^+, r_2^+, \ldots, r_{N^+}^+, r_1^-, r_2^-, \ldots, r_{N^-}^- \ge 0$ of \emph{radii} satisfying
\begin{enumerate}
\item $\bigcup_{i=1}^{N^+} X^+_i$ and $c_i^+\in\conv (X^+_i)$ where $X^+_i := X^+\cap \B_2^d(c_{i}^+, r_{i}^+)$,
\item $\bigcup_{j=1}^{N^-} X^-_j$ and $c_j^-\in\conv (X^-_j)$ where $X^-_j := X^-\cap \B_2^d(c_{j}^-, r_{j}^-)$,
\item $r_i^+ \le \mu^{-1}\dist^{2}\left(c_i^+, C^{-}\right)$,
\item $r_j^- \le \mu^{-1}\dist^{2}\left(c_j^-, C^{+}\right)$,
\item and $C^{+}, C^{-}$ are $\delta$-separated
\end{enumerate}
for every $i\in [N^+]$ and $j\in [N^-]$. As a special case where $\mu\to \infty$, the \emph{mutual complexity} of two finite sets $X^{+}$ and $X^{-}$ is the cardinality $N=\vert X^+\vert +\vert X^-\vert$. 

\end{definition}

\subsection{Main Results}
In this section, we present our main results. Our first major result, denoted as \cref{thm:main}, encompasses two key aspects. The initial component demonstrates the capability of a one-layer RINN to render any two finite $\delta$-separated sets linearly separable. The subsequent component extends this outcome to two arbitrary $\delta$-separated sets with a predefined mutual complexity $N$. Both aspects of our first major result are established using two distinct weight distributions for the random initialization of the neural network: one samples i.i.d. copies from $\S^{d-1}$ as rows, while the other employs a standard Gaussian matrix.

\begin{theorem}\label{thm:main}
Fix any $\eta \in (0, 1)$ and $\delta$-separated $X^{+}, X^{-}\subset R\B_{2}^{d}$ with $d\geq 2$.
\begin{itemize}
	\item If $X^+$ and $X^-$ are finite, let $N:=\vert X^{+}\vert +\vert X^{-}\vert$, and let $\gamma: =\delta^{2}/8Rd$.
	\item Otherwise, let $N$ be the $(\delta, 64\delta^{-2}R^{3}d)$-mutual complexity of $X^+$ and $X^-$, and let $\gamma:= \delta^2/16Rd$.
\end{itemize}
For a one-layer RINN $\Phi(x):=\Re(Wx+b)$ with weights $W\in\R^{n\times d}$ and bias $b\sim \Unif\left([-\lambda, \lambda]^n\right)$, the sets $\Phi(X^{+}), \Phi (X^{-})\subset\R^{n}$ are linearly separable with probability at least $1-\eta$ by a margin at least $\mathcal{M}(\gamma,N)$ if width $n$ and maximal bias $\lambda$ satisfy either of the following, depending on how weights $W$ is sampled:
\begin{enumerate}
\item[(i)] $n\geq \frac{\log (N/\eta)}{p}$ and $\lambda\geq R$ if each row of the weight matrix $W_i\iid \Unif(\S^{d-1})$,
\item[(ii)] $n\geq \frac{10\log (N/\eta)}{p}$ and $\lambda\geq 3R\sqrt{d}$ if weights $W\sim \N(0, I_{n\times d})$ are Gaussian,
\end{enumerate}
where we recall that $\N(0, I_{n\times d})$ means each entry is i.i.d. $\N(0, 1)$, and define
\begin{equation}\label{eq:p}
p:=\frac{R}{8(d-1)\lambda}\left(\frac{\delta^{2}}{8R^{2}}\right)^{d}\quad\text{and}\quad\mathcal{M}(\gamma,N):=\sqrt{\frac{4R(R+\gamma)}{N(1+2R/\gamma)^{2N}-N}}.
\end{equation} 
\end{theorem}

\cref{thm:main} shows for the first time that a $\Re$-activated one-layer RINN transforms any two $\delta$-separated sets into two linearly separable sets with high probability, where the required width is polynomial in the parameters of the random weight distributions and $\delta^{-1}$. This advances our understanding on internal geometry of neural network as it improves further the best known bound given by  \cite[Theorem~2.3]{dirksen2021separation} where the authors showed that required width grows exponentially in those variables for linear  separation of two given sets of points. 

\cref{thm:main} can be extended to show the separation capacity of $\Phi$ for $\ell$-many $\delta$-separated sets by running the argument between each set and the union of the remaining sets. Let $N$ be the total number of points if the sets are finite, or the $(\delta, 64\delta^{-2}R^3d)$-mutual complexity of all sets otherwise. For any $\eta\in (0, 1)$, with $\ell$ times the width lower bound, $\Phi$ makes each set linearly separable from the rest with the same margin and probability at least $1-\ell\eta$ by a union bound.

The following result extends the scope of \cref{thm:main} by improving the bound on the width of the one-layer RINN. Unlike the exponential dependence of width on dimension in \cref{thm:main}, the proposed width in our next result will only grow polynomially in the dimension $d$ of the data. To state it, we define the \emph{Gaussian width} $w(T)$ of a subset $T\subset \mathbb{R}^n$ (\cite[Definition~7.5.1]{Ver18}) as
\begin{equation}\label{eq:w}
 w(T) := \mathbb{E}\left[\sup_{x\in T} g^{\intercal}x\right] \text{ where } g\sim \N(0, I_{n}).
\end{equation}

\begin{theorem}\label{thm:main in k}
There exists a constant $C = C(\delta, R)>0$ such that the following holds. Fix any $\eta \in (0, 1)$ and $\delta$-separated $X^{+}, X^{-}\subset R\B_{2}^{d}$ with $d\geq 2$.

\begin{itemize}
	\item If $X^+$ and $X^-$ are finite, let $N:=\vert X^{+}\vert +\vert X^{-}\vert$, and let $\gamma: =\delta^{2}/18R\sqrt{k}$.
	\item Otherwise, let $N$ be the $(\delta, 144\delta^{-2}R^{3}\sqrt{k})$-mutual complexity of $X^+$ and $X^-$, and let $\gamma:= \delta^{2}/36R\sqrt{k}$, where in both cases we let
\begin{equation}\label{eq:k}
 k := C\left(\frac{32R}{\delta^2}\right)^2\left(w\left((X^+-X^-)\cup X^+\cup X^-\right)^2+R^2\right).
\end{equation}
\end{itemize}
For a one-layer RINN $\Phi(x):=\Re(Wx+b)$ with standard Gaussian weights $W\sim \N(0, I_{n\times d})$ and bias $b\sim \Unif\left([-\lambda, \lambda]^n\right)$, the sets $\Phi(X^{+}), \Phi (X^{-})\subset\R^{n}$ are linearly separable with probability at least $1-\eta$ by a margin at least $\mathcal{M}(\gamma,N)$ defined in \eqref{eq:p}, provided maximal bias $\lambda \geq 9R\sqrt{k}/8$ and width \begin{equation}\label{eq:q}
n \geq \frac{\log (N/\eta)}{q}, \quad \text{where}\quad  q :=\frac{R}{4\lambda \sqrt{k}}\left(\frac{2\delta^2}{81R^2}\right)^k.
\end{equation}
\end{theorem}

\cref{thm:main in k} offers an alternative bound to address the same problem tackled in \cref{thm:main}. This result demonstrates that the width of a one-layer RINN does not necessitate exponential growth concerning dimensionality, thereby providing a solution to the issue of the curse of dimensionality.

To distill the essence of both \cref{thm:main,thm:main in k}, we summarize as follows. When dealing with data of constant dimensionality, \cref{thm:main} indicates that merely $O(\log n)$ neurons within a one-layer RINN are required to achieve linear separability with a probability of at least $1-1/n$. However, when the data intrinsic dimension of data is uncertain, \cref{thm:main in k} offers a solution: it ensures linear separation with a probability of at least $1-1/n$ by employing a one-layer RINN consisting of a mere $O(n^c\log n)$ neurons, where the value of $c$ is explicitly defined in terms of the initial data separation.

\subsection{Related Works}\label{sec:relworks}
Expressivity of NNs has been the focus of the literature on NNs since the 1960s. Earlier works along this direction study the memorization capacity of linear threshold networks \cite{baum1988capabilities,cover1965geometrical,kowalczyk1997estimates,sontag1997shattering}. Later, their counterparts with sigmoids, $\Re$, and other bounded activation functions have been studied \cite{huang1991perceptrons,yamasaki1993lowercap}. More works show the empirical evidence of training and performance advantages \cite{nair2010relu,Krizhevsky2012alexnet,maas2013relu,glorot2011deepsparse,dahl2013LVCSR,sun2014facerep}. While the amount of empirical evidence is highly impressive, only few provided theoretical justification for such performance. Most theoretical work focuses on topics such as universal approximation power \cite{sun2018approximation, andoni2014learning, yehudai2019power,needell2020random,hsu2021approximation}, finite sample expressivity, or memorization capacity of NNs \cite{zhang2016understanding,hardt2016identity,nguyen2018optimization,yun2019small}, none of which guarantees that NNs perform well in data separation.

The separation capacity of NNs has recently been explored in \cite{an2015can}, where the authors put forth a systematic geometric framework for showing that any two disjoint subsets $X^{+}, X^{-}\subset \mathbb{R}^d$ can be linearly separated using a deterministic two-layer NN. Their construction has very recently been adopted for RINNs by \cite{dirksen2021separation} to study the separation capacity of two-layer RINNs, with asymptotic bounds on the output dimension, unlike the existential results in \cite{an2015can}. Our results show that linear separability of the data can actually be achieved with a one-layer RINN without the imposition of the curse of dimensionality. 

Kernel-based methods are a major tool for linear separation of the data. Introduced by \cite{rahimi2007random}, random feature methods provide a cheap alternative to the kernel evaluation of the data via the construction of random feature maps in high dimensions. The generalization error of random feature maps is further studied in \cite{rahimi2007random,rahimi2008weighted,bach2017equivalence,rudi2017generalization} and is shown to be comparable to the kernel-based methods. The idea of using RINNs \cite{an2015can,dirksen2021separation} to separate data shares many similarities with the use of random feature methods. However, our results have no direct implication for the theory of random feature maps.

{The generalization properties of estimation with random features
in the statistical learning framework has been studied in depth. One of the best possible results is obtain by \cite{rudi2017generalization}, where they showed that $O(1/\sqrt{n})$ learning bounds can be achieved with only $O(\sqrt{n}\log n)$ random features in the kernelized ridge regression setting rather than $O(n)$
as suggested by previous results. Furthermore, they have shown faster learning rates and show
that they might require more random features, unless they are sampled according to
a possibly problem dependent distribution. However this type of result does not shed much light on the number of features needed in the data separation problem.

There had been immense progress in perfectly separating data sets using overparameterized neural networks trained by gradient descent under the NTK (neural tangent kernel) regime. 
One of the significant results was obtained by \cite{allen-zhu19GlobalConvergence} which showed that if the task is multi-label classification, then gradient descent/stochastic gradient descent finds a 100\% accurate classifier on the
training set for feedforward neural networks with ReLU activation if each layer is sufficiently wide. Most of those works, including \cite{jacot2018NTK,allen-zhu19GlobalConvergence, LiLiang2018learningoverparametrized, yun2019small} have assumed that the network is sufficiently wide. In \cite{Gao2019} determined the minimum number of hidden neurons required to attain a specified level of adversarial training loss. \cite{Daniely2020} demonstrated that a depth-two neural network can memorize labels for a dataset consisting of $m$ samples of $d$-dimensional vectors using $O(m/d)$ hidden neurons. \cite{Ji2020Polylogarithmic} showcased that a depth-two neural network can achieve nearly 100\% accuracy with a poly-logarithmic number of hidden neurons relative to the number of samples. They also established bounds on generalization error, making their results a state-of-the-art contribution to the field.
\cite{Oymak2019TowardMO} demonstrated that deep neural networks require bounds on hidden neurons and gradient descent steps that grow polynomially to achieve a desired level of accuracy.
However, our paper aims to quantify how much width is sufficient for a randomly initialized one-layer neural network to make data linearly separable without any training. 
 
Compared to learning with over-parameterized networks, our result has one significant advantage. Our bound on the width on the network is explicitly written in terms of sample size, initial separation of the data sets and the separation probability. \cref{thm:main} shows that we just need $O(\log n)$ neurons for a one-layer RINN to make data linearly separable with probability $1-1/n$ for fixed dimension. If the intrinsic dimension of the data is unknown, \cref{thm:main in k} guarantees linear separation with probability $1-1/n$ using a one-layer RINN with just $O(n^c\log n)$ neurons, where $c$ is made explicit.
} 

A key point in our analysis is distance preservation by ReLU-activated RINNs. This feature, first sketched in \cite{an2015can} and shown rigorously in \cite{dirksen2021separation}, is reminiscent of nonlinear random embedding in compressed sensing \cite{jacques2013robust,plan2014dimension,oymak2015near,dirksen2021non}.

We end this section by discussing similarities between our results and the rare eclipse problem. For any two linearly separable sets, the rare eclipse problem searches for a random lower-dimensional embedding that makes them disjoint with high probability. It was shown in \cite{bandeira2017compressive} via Gordon's escape through the mesh theorem \cite{gordon1988milman} that multiplication by a random Gaussian matrix achieves such an embedding with a sharp lower bound on the dimension. This bound in \cite[Corollary~3.2]{bandeira2017compressive}, stated in terms of the Gaussian width of the difference between two sets, shares close ties with the expression of $k$ in \eqref{eq:k} of \cref{thm:main}.

\subsection{Proof Strategy}\label{sec:sketch}
Our primary focus lies on providing a proof sketch when dealing with finite sets $X^{+}$ and $X^{-}$. Extending this proof to arbitrary sets is fairly straightforward. To maintain brevity, we will present a brief sketch of the argument for separation while omitting the margin of guarantee. Additionally, we will freely switch between the uniform unit row vector and the standard Gaussian initializations of the weight matrix $W$ in \cref{thm:main}, with a brief comment on their similarities. 

The following interpretation of a one-layer neural network $\Phi$ is central to our argument: recall that $i$-th node of $\Phi$ encodes the $i$-th row $W_i$ of the weight matrix $W$ and the $i$-th entry $b_i$ of $b$, i.e., $\left(\Phi(x)\right)_{i} = \Re\left(W_{i}^{\intercal}x+b_{i}\right)$. For any $x\in \mathbb{R}^n$, assuming that $W_i\in\S^{d-1}$ is normalized, $h_i(x)=W^{\intercal}_ix+b_i$ is the signed distance between $x$ and the hyperplane $\lbrace h_i=0\rbrace$. Thus, $(\Phi(x))_{i}$ measures the distance of $x$ and $h_i$ if $x$ is on the positive side of $h_i$ and is zero otherwise.

\paragraph{Step 1: Constructing a Deterministic Layer $\Phi$}
The first step, outlined in \cref{sec:One}, is the construction of a deterministic one-layer NN that makes $X^+$ and $X^-$ linearly separable, if they are both finite. The construction is given in \cref{alg:det_onelayer}: we enumerate all points $x_i\in X^+\cup X^-$ in descending order of norm. For each $x_i$, we will find a hyperplane $h_i$ where $x_i$ is on the positive side of $h_i$ and all $x_j$ in the opposite class with $j>i$ (i.e. $\Vert x_j\Vert \le\Vert x_i\Vert$ are on the negative side of $h_i$. This condition is formalized as $B_i$ in \cref{thm:prob}. 

\paragraph{Step 2: Showing $\Phi(X^+)$ and $\Phi(X^-)$ are Linearly Separable}

We find the separating hyperplane via \cref{alg:weights}. This amounts to finding a vector $a$ such that $a^{\intercal}\Phi(x^{+}) >0$ for every $x^{+}\in X^{+}$ and $a^{\intercal}\Phi(x^{-}) <0$ for every $x^{-}\in X^{-}$.  Using the above guarantee, we will construct $a$ coordinate-by-coordinate recursively, with $a_i$ chosen so that $a$ separates $\Phi(X^{+})$ and $\Phi (X^{-})$.

\paragraph{Step 3: Passing to a Randomly Initialized Layer}
Under random initialization, we need two additional ideas: first, by \cref{lem:proj}, if a layer makes the sets linearly separable, it continues to do so after adding more nodes. The second idea, shown in \cref{thm:prob}, is to argue that the event $B_i$ occurs with positive probability for a randomly chosen hyperplane $h_i$. 
Combining these ideas, when the width is sufficiently large, we can find for every $i$ a copy of $h_i$ satisfying $B_i$ in the random layer. These together makes $X^+$ and $X^-$ linearly separable.

\paragraph{Step 4: Generalizing to Arbitrary Sets}
To obtain the result for arbitrary sets, we use the notion of mutual complexity in \cref{def:complexity} which is influenced by \cite{dirksen2021separation}. Roughly, the idea is to cover the sets with disjoint balls of small radius, and run the argument on the finite sets of centers $C^+$ and $C^-$. Upon choosing parameters carefully, we will show the linear separability of $\Phi(C^+)$ and $\Phi(C^-)$ implies that of $\Phi(X^+)$ and $\Phi(X^-)$ , with some margins, as in \cref{thm:arb}.

\paragraph{Step 5: Uniform versus Gaussian Random Weight Matrix} We briefly comment on the two choices of random initialization of our weight matrices. While our analysis seems to be tailored for weight matrix $W$ with rows being uniform random unit vectors, and we use simple results like \cref{lem:prob lm} and \cref{prop:chi} to translate into the Gaussian case, the Gaussian case is far from an afterthought. In particular, in \cref{thm:prob}(iii), we factor a Gaussian vector into a Gaussian matrix and a uniform random unit vector, as in \cite[Proof of Theorem 1.6]{dirksen2021separation}, and apply a matrix deviation inequality in \cite[Exercise 9.1.8]{Ver18} to get a bound independent of $d$, thereby escaping the curse of dimensionality.

Below we present a schematic representation of the proof ideas using a flow chart.

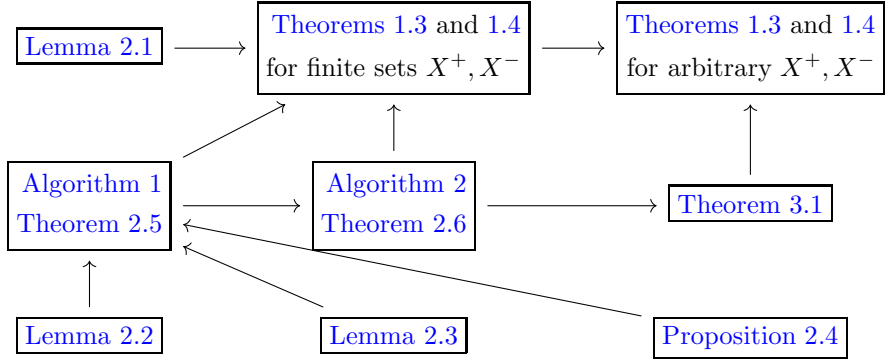
\[\begin{tikzcd}
\boxed{\text{\cref{lem:proj}}} \arrow[r]                            & \boxed{\begin{aligned}\text{\cref{thm:main,thm:main in k}} \\ \text{for finite sets $X^+, X^-$}\end{aligned}} \arrow[r] & \boxed{\begin{aligned}\text{\cref{thm:main,thm:main in k}} \\ \text{for arbitrary $X^+, X^-$}\end{aligned}} \\
\boxed{\begin{aligned}\text{\cref{alg:det_onelayer}}\\ \text{\cref{thm:prob}}\end{aligned}} \arrow[ru] \arrow[r] & \boxed{\begin{aligned}\text{\cref{alg:weights}}\\\text{\cref{thm:fin}}\end{aligned}} \arrow[u] \arrow[r]          & \boxed{\text{\cref{thm:arb}}} \arrow[u]                             \\
\boxed{\text{\cref{lem:geo lm}}} \arrow[u]                          & \boxed{\text{\cref{lem:prob lm}}} \arrow[lu]           & \boxed{\text{\cref{prop:chi}}} \arrow[llu]  
\end{tikzcd}\]

\section{A One-Layer Deterministic Neural Network Separates Finite Sets}\label{sec:One}
In this section, we sketch and prove the deterministic NN analogs of the finite set cases of \cref{thm:main}, namely steps 1 and 2 in the proof sketch.

\subsection{Some Preliminary Lemmas}\label{sec:prelim}
Before we begin, we will state and prove a few results in convex geometry and probability. As outlined in the proof strategy section, the following simple observation is key for us to go from a deterministic layer to a random layer.

\begin{lemma}\label{lem:proj}
For any $S^{+}, S^{-}\subset \R^{k}$ and projection $\pi$ onto a subspace indexed by coordinates $\Sigma \subset \lbrace 1, \dots , k\rbrace$, if $\pi(S^{+})$ and $\pi (S^{-})$ are linearly separable with margin $\mu$, then $S^{+}$ and $S^-$ are also linearly separable with margin at least $\mu$.
\end{lemma}
\begin{proof}
Let $\Sigma = \lbrace \sigma_{1}, \dots , \sigma_{\ell} \rbrace$. If $\pi(S^{+})$ and $\pi (S^{-})$ are linearly separable, then there exists $a=(a_{\sigma(x_i) })_{i}\in \R^{\ell}$ with $\Vert a\Vert =1$ and $b\in \mathbb{R},\mu\in \mathbb{R}_{\geq 0}$ such that 
 the hyperplane defined by $s\mapsto a^\intercal s+b$ separates $\pi(S^+)$ and $\pi(S^-)$ with margin $\mu$. 
Define $\tilde{a}\in\R^{k}$ by $\tilde{a}_i = a_i\mathbf{1}\{i\in \Sigma\}$ for $i\in [k]$. Then
\[ \tilde{a}^{\intercal}s +b = \sum_{j=1}^{k}a_{j}s_{j}+b=\sum_{i=1}^{\ell} a_{\sigma(x_i) }s_{\sigma(x_i) } +b =a^{\intercal}\pi (s) +b = \begin{cases}
\geq \mu & \textup{if }s\in S^{+}\\
\leq -\mu & \textup{if }s\in S^{-}
\end{cases}.\]
As $\Vert \tilde{a}\Vert \le \Vert a\Vert =1$, $S^+$ and $S^-$ are linearly separable with margin at least $\mu$.
\end{proof}

\begin{lemma}\label{lem:geo lm}
Fix $d \in \mathbb{Z}_{\geq 2}$. Denote the convex hull of $X$ by $\conv(X)$.
\begin{enumerate}
\item For any $0\leq r\leq 2$, $u\in \S^{d-1}$, and $v\sim\Unif (\S^{d-1})$, we have \[\P_{v} \left(\Vert u-v\Vert \leq r\right)\geq \frac{1}{2}\left(\frac{r}{2}\right)^{d-1}.\]
\item For any set $X\subset \R^d$ and $c\in\conv(X)$, there exists $x\in X$ with $\Vert x\Vert\geq \Vert c\Vert$. Moreover, for any vector $g\in\R^d$, there exists some $x\in X$ with $g^{\intercal}x\geq g^{\intercal}c$.
\end{enumerate}
\end{lemma}
\begin{proof} 
(1) follows from \cite[Lemma 2.3]{ball1997elementary}. For (2), note that since $c\in \conv (X)$, there are weights $\lbrace w_x:x\in X\rbrace\subset [0, 1]$ such that
\[ c=\sum _{x\in X}w_xx\quad\text{and}\quad \sum_{x\in X}w_x=1.\]
By the triangle inequality and linearity of inner product with $g$
\[ \Vert c\Vert \leq \sum_{x\in X}w_x\Vert x\Vert \quad\text{and}\quad g^{\intercal}c=\sum _{x\in X}w_xg^{\intercal}x.\]
As $w_x$ are non-negative and $\sum_{x\in X}w_x= 1$, (2) follows the equation above. 
\end{proof}
\begin{lemma}\label{lem:prob lm}
Let $z\sim \chi_{d}^{2}$ be the chi-squared distributions with parameter $d$.
\begin{enumerate}
\item For $v\sim \N(0, I_{d})$, $\rho := \Vert v\Vert $ satisfies $\rho^2\sim \chi^2_{d}$, and $\hat{v}:=v/\rho\sim \Unif(\S^{d-1})$. Moreover, $\rho$ and $\hat{v}$ are independent.
\item$\P(z-d\geq 2\sqrt{dx}+2x)\leq e^{-x}$ and $\P(d-z\geq 2\sqrt{dx})\leq e^{-x}$ for all $x>0$.
\end{enumerate}
\end{lemma}
\begin{proof}
(1) is well-known. (2) follows from \cite[Lemma 1]{laurent2000adaptive}.
\end{proof}
\begin{proposition}\label{prop:chi}
Suppose $\rho^2\sim \chi^2_d$, then $\P(1\leq \rho\leq 3\sqrt{d})\geq 1/10$.
\end{proposition} 
\begin{proof} 
By the second inequality of \cref{lem:prob lm}(2) with $x=(d-1)^2/4d$ \[ \P(\rho\leq 1)=\P(d-\rho^2\geq d-1)\leq e^{-(d-1)^2/4d}\] which is decreasing for $d\geq 2$, so it attains its maximum at $d=2$. Hence
\[\P(\rho\leq 1)\leq e^{-(d-1)^2/4d} \leq e^{-1/8}.\]
Now, let $x=9/2$. We see that $9d\geq d+\sqrt{2x}d+dx$ for $d\geq 2$, so
\begin{align*}
\P(\rho \geq 3\sqrt{d})  &= \P(\rho^2 \geq 9d)
\leq \P(\rho^2\geq d+\sqrt{2x}d+dx)
\\ & \leq \P(\rho^2\geq d+2\sqrt{dx}+2x)
\leq \P(\rho^2-d\geq 2\sqrt{dx}+2x)
\leq e^{-9/2}.
\end{align*}
Finally, by a union bound, $\P(1\leq\rho\leq 3\sqrt{d})\geq 1-e^{-1/8}-e^{-9/2}\geq 1/10$.
\end{proof}

\subsection{Constructing the Deterministic Layer $\Phi$}

We use \cref{alg:det_onelayer} below to construct the one-layer deterministic NN by drawing the hyperplanes separating one point at a time in descending order of norm. Before giving \cref{alg:det_onelayer}, we introduce some notations. Recall $k$ from \eqref{eq:k}. Let $N:=\vert X^{+}\vert +\vert X^{-}\vert$ and define \emph{sign} $\sigma: X^{+}\cup X^{-}\to \{+1,-1\}$ by 
\begin{equation}\label{eq:sigma}
\sigma(x)=+1 \quad\text{  if  }\quad  x\in X^{+}, \quad \text{and} \quad  \sigma(x)=-1 \quad \text{  if  }\quad  x\in X^{-}.  
\end{equation}
For convenience of visualization, we will often refer to $h_i$ (see line 6 of \cref{alg:det_onelayer}) as the hyperplane $\lbrace h_i=0\rbrace$. The goal of \cref{alg:det_onelayer} is to construct the hyperplane $h_i$ which separates $x_{i}$ by a margin $\gamma$ from all points $x_j\in X^{+}\cup X^{-}$ which have smaller norm than $x_i$ and $\sigma(x_j)\neq \sigma(x_i)$.
\begin{algorithm}[H]
\caption{Construct a one-layer NN separating two finite sets with a margin}
\label{alg:det_onelayer}
\begin{algorithmic}[1]
\Require Two $\delta$-separated finite sets $X^{+}, X^{-}\subset R\B_{2}^{d}$ and $\gamma \in [0, \max(\delta^2/8Rd, \delta^2/18R\sqrt{k})]$.
\Ensure A one-layer neural network $\Phi$ such that $\Phi(X^{+})$ and $\Phi(X^{-})$ are linearly separable.

\For{$i = 1,2,\ldots,N$}
\State $x_i \gets x$ where $x$ has the $i$-th largest norm in $X^+ \cup X^-$.
\EndFor
\For{$i = 1,2,\ldots,N$}

    \State Find $w\in \S^{d-1}$ and $b\in \mathbb{R}$ such that $w^{\intercal}x_i +b\geq \gamma$ and $w^{\intercal}x_j+b\leq -\gamma$ for all $j > i$ and $\sigma(x_i) \neq \sigma(x_j)$. If no such $j$ exists, ask $w^{\intercal}x_i +b\geq \gamma$ and $w^{\intercal}x +b \leq 0$ for some $x\in R\B_2^d$.
    \State $w_i \gets w$, $b_i \gets b$, $h_i \gets$ the function $w^{\intercal}_ix + b_i$ %
\EndFor
\State $W \gets$ the matrix with row vectors $w_i^{\intercal}$
\State $b \gets$ the vector with components $b_i$
\State \Return the function $\Phi(x)=\Re(Wx + b)$
\end{algorithmic}
\end{algorithm}

It is easy to see that each hyperplane $h_i$ exists in the case where no such $j$ exists on line 5. However, it is not obvious why $h_i$ exists when $j$ exists. In the next result, we will show this via the probabilistic method that $h_i$ exists when $\gamma$ is at most either term in the maxima in the input condition.
\begin{theorem}\label{thm:prob}
For any two $\delta$-separated finite sets $X^{+}, X^{-}\subset R\B_{2}^{d}$, define $x_i$ and $\sigma$ as in \cref{alg:det_onelayer} and \eqref{eq:sigma}. Consider the random affine function $h(z)=v^{\intercal}z+t$ where $t\in \mathbb{R},v\in \mathbb{R}^d$ are independent random variables with $t\sim\Unif\left([-\lambda, \lambda]\right)$ and the distribution of $v$ is specified below. For any $\gamma>0$, let
\[ B_{i}:=\left\lbrace h(x_{i})\geq \gamma \right\rbrace\cap \bigcap_{j:j>i, \sigma(x_i) \neq\sigma(x_j)  }\left\lbrace h(x_{j})\leq -\gamma \right\rbrace\]
for each $i\in [N]$. Recall $p$ from \eqref{eq:p} and $q$ from \eqref{eq:q}. Then, there exists some $C=C(\delta, R)>0$ in the definition \eqref{eq:k} of $k$ such that the following hold:
\begin{enumerate}
    \item[(i)] For all $i$ and $\gamma \leq {\delta^{2}}/{8Rd}$, $\P (B_{i})\geq p$ if $v\sim\Unif \left(\S^{d-1}\right)$ and $\lambda\geq R$.
    \item[(ii)] For all $i$ and $\gamma \leq {\delta^{2}}/{8Rd}$, $\P (B_{i})\geq p/10$ if $v\sim \N(0, I_{d})$ and $\lambda\geq 3R\sqrt{d}$.
    \item[(iii)] For all $i$ and $\gamma \leq {\delta^{2}}/{18R\sqrt{k}}$, $\P (B_{i})\geq q$ if $v\sim \N(0, I_{d})$ and $\lambda\geq 9R\sqrt{k}/8$.
\end{enumerate}
\end{theorem}

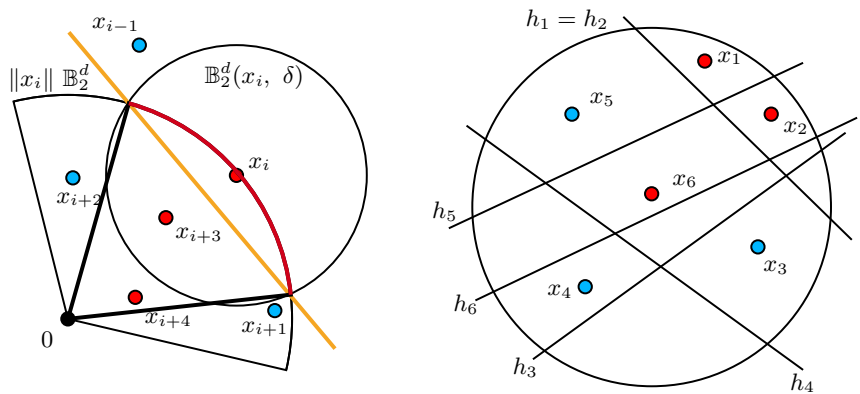
\begin{figure}[H]
    \centering

\tikzset{every picture/.style={line width=0.75pt}} 

\begin{tikzpicture}[x=0.75pt,y=0.75pt,yscale=-0.67,xscale=0.67]

\draw   (377.67,157.67) .. controls (377.67,83.11) and (438.11,22.67) .. (512.67,22.67) .. controls (587.23,22.67) and (647.67,83.11) .. (647.67,157.67) .. controls (647.67,232.23) and (587.23,292.67) .. (512.67,292.67) .. controls (438.11,292.67) and (377.67,232.23) .. (377.67,157.67) -- cycle ;
\draw  [draw opacity=0][fill={rgb, 255:red, 255; green, 0; blue, 0 }  ,fill opacity=1 ] (547.67,47.67) .. controls (547.67,44.91) and (549.91,42.67) .. (552.67,42.67) .. controls (555.43,42.67) and (557.67,44.91) .. (557.67,47.67) .. controls (557.67,50.43) and (555.43,52.67) .. (552.67,52.67) .. controls (549.91,52.67) and (547.67,50.43) .. (547.67,47.67) -- cycle ;
\draw  [draw opacity=0][fill={rgb, 255:red, 0; green, 183; blue, 255 }  ,fill opacity=1 ] (447.67,87.67) .. controls (447.67,84.91) and (449.91,82.67) .. (452.67,82.67) .. controls (455.43,82.67) and (457.67,84.91) .. (457.67,87.67) .. controls (457.67,90.43) and (455.43,92.67) .. (452.67,92.67) .. controls (449.91,92.67) and (447.67,90.43) .. (447.67,87.67) -- cycle ;
\draw  [draw opacity=0][fill={rgb, 255:red, 0; green, 183; blue, 255 }  ,fill opacity=1 ] (587.67,187.67) .. controls (587.67,184.91) and (589.91,182.67) .. (592.67,182.67) .. controls (595.43,182.67) and (597.67,184.91) .. (597.67,187.67) .. controls (597.67,190.43) and (595.43,192.67) .. (592.67,192.67) .. controls (589.91,192.67) and (587.67,190.43) .. (587.67,187.67) -- cycle ;
\draw  [draw opacity=0][fill={rgb, 255:red, 0; green, 183; blue, 255 }  ,fill opacity=1 ] (457.67,217.67) .. controls (457.67,214.91) and (459.91,212.67) .. (462.67,212.67) .. controls (465.43,212.67) and (467.67,214.91) .. (467.67,217.67) .. controls (467.67,220.43) and (465.43,222.67) .. (462.67,222.67) .. controls (459.91,222.67) and (457.67,220.43) .. (457.67,217.67) -- cycle ;
\draw  [draw opacity=0][fill={rgb, 255:red, 255; green, 0; blue, 0 }  ,fill opacity=1 ] (597.67,87.67) .. controls (597.67,84.91) and (599.91,82.67) .. (602.67,82.67) .. controls (605.43,82.67) and (607.67,84.91) .. (607.67,87.67) .. controls (607.67,90.43) and (605.43,92.67) .. (602.67,92.67) .. controls (599.91,92.67) and (597.67,90.43) .. (597.67,87.67) -- cycle ;
\draw  [draw opacity=0][fill={rgb, 255:red, 255; green, 0; blue, 0 }  ,fill opacity=1 ] (507.67,147.67) .. controls (507.67,144.91) and (509.91,142.67) .. (512.67,142.67) .. controls (515.43,142.67) and (517.67,144.91) .. (517.67,147.67) .. controls (517.67,150.43) and (515.43,152.67) .. (512.67,152.67) .. controls (509.91,152.67) and (507.67,150.43) .. (507.67,147.67) -- cycle ;
\draw    (491.33,14.33) -- (663.33,182.33) ;
\draw    (360,173.67) -- (626.67,49) ;
\draw    (373.33,96.33) -- (626.67,280.33) ;
\draw    (423.33,272.33) -- (658.67,101.67) ;
\draw  [draw opacity=0][fill={rgb, 255:red, 0; green, 0; blue, 0 }  ,fill opacity=1 ] (68.25,241.92) .. controls (68.25,239.16) and (70.49,236.92) .. (73.25,236.92) .. controls (76.01,236.92) and (78.25,239.16) .. (78.25,241.92) .. controls (78.25,244.68) and (76.01,246.92) .. (73.25,246.92) .. controls (70.49,246.92) and (68.25,244.68) .. (68.25,241.92) -- cycle ;
\draw  [draw opacity=0][fill={rgb, 255:red, 255; green, 0; blue, 0 }  ,fill opacity=1 ] (195.13,133.79) .. controls (195.13,131.03) and (197.36,128.79) .. (200.13,128.79) .. controls (202.89,128.79) and (205.13,131.03) .. (205.13,133.79) .. controls (205.13,136.55) and (202.89,138.79) .. (200.13,138.79) .. controls (197.36,138.79) and (195.13,136.55) .. (195.13,133.79) -- cycle ;
\draw  [draw opacity=0] (32.3,78.17) .. controls (45.41,74.9) and (59.13,73.17) .. (73.25,73.17) .. controls (166.45,73.17) and (242,148.72) .. (242,241.92) .. controls (242,255.2) and (240.47,268.12) .. (237.56,280.52) -- (73.25,241.92) -- cycle ; \draw   (32.3,78.17) .. controls (45.41,74.9) and (59.13,73.17) .. (73.25,73.17) .. controls (166.45,73.17) and (242,148.72) .. (242,241.92) .. controls (242,255.2) and (240.47,268.12) .. (237.56,280.52) ;  
\draw   (102,133.79) .. controls (102,79.6) and (145.93,35.67) .. (200.13,35.67) .. controls (254.32,35.67) and (298.25,79.6) .. (298.25,133.79) .. controls (298.25,187.98) and (254.32,231.92) .. (200.13,231.92) .. controls (145.93,231.92) and (102,187.98) .. (102,133.79) -- cycle ;
\draw [color={rgb, 255:red, 245; green, 166; blue, 35 }  ,draw opacity=1 ][line width=1.5]    (74,26) -- (229.3,210.75) -- (274.33,264.33) ;
\draw  [draw opacity=0][fill={rgb, 255:red, 0; green, 183; blue, 255 }  ,fill opacity=1 ] (224,235.67) .. controls (224,232.91) and (226.24,230.67) .. (229,230.67) .. controls (231.76,230.67) and (234,232.91) .. (234,235.67) .. controls (234,238.43) and (231.76,240.67) .. (229,240.67) .. controls (226.24,240.67) and (224,238.43) .. (224,235.67) -- cycle ;
\draw  [draw opacity=0][fill={rgb, 255:red, 0; green, 183; blue, 255 }  ,fill opacity=1 ] (72,135.67) .. controls (72,132.91) and (74.24,130.67) .. (77,130.67) .. controls (79.76,130.67) and (82,132.91) .. (82,135.67) .. controls (82,138.43) and (79.76,140.67) .. (77,140.67) .. controls (74.24,140.67) and (72,138.43) .. (72,135.67) -- cycle ;
\draw  [draw opacity=0][fill={rgb, 255:red, 255; green, 0; blue, 0 }  ,fill opacity=1 ] (142,165.67) .. controls (142,162.91) and (144.24,160.67) .. (147,160.67) .. controls (149.76,160.67) and (152,162.91) .. (152,165.67) .. controls (152,168.43) and (149.76,170.67) .. (147,170.67) .. controls (144.24,170.67) and (142,168.43) .. (142,165.67) -- cycle ;
\draw  [draw opacity=0][fill={rgb, 255:red, 0; green, 183; blue, 255 }  ,fill opacity=1 ] (122,35.67) .. controls (122,32.91) and (124.24,30.67) .. (127,30.67) .. controls (129.76,30.67) and (132,32.91) .. (132,35.67) .. controls (132,38.43) and (129.76,40.67) .. (127,40.67) .. controls (124.24,40.67) and (122,38.43) .. (122,35.67) -- cycle ;
\draw  [draw opacity=0][fill={rgb, 255:red, 255; green, 0; blue, 0 }  ,fill opacity=1 ] (119,225.67) .. controls (119,222.91) and (121.24,220.67) .. (124,220.67) .. controls (126.76,220.67) and (129,222.91) .. (129,225.67) .. controls (129,228.43) and (126.76,230.67) .. (124,230.67) .. controls (121.24,230.67) and (119,228.43) .. (119,225.67) -- cycle ;
\draw  [draw opacity=0][line width=1.5]  (119.15,79.48) .. controls (184.39,97.88) and (233.59,154.52) .. (241.02,223.66) -- (73.25,241.92) -- cycle ; \draw  [color={rgb, 255:red, 208; green, 2; blue, 27 }  ,draw opacity=1 ][line width=1.5]  (119.15,79.48) .. controls (184.39,97.88) and (233.59,154.52) .. (241.02,223.66) ;  
\draw    (667.33,90.33) -- (380,228) ;

\draw (51,249.07) node [anchor=north west][inner sep=0.75pt]  [font=\small]  {$0$};
\draw (207,118.07) node [anchor=north west][inner sep=0.75pt]  [font=\small]  {$x_{i}$};
\draw (152,171.07) node [anchor=north west][inner sep=0.75pt]  [font=\small]  {$x_{i+3}$};
\draw (62.33,144.4) node [anchor=north west][inner sep=0.75pt]  [font=\small]  {$x_{i+2}$};
\draw (89.33,13.07) node [anchor=north west][inner sep=0.75pt]  [font=\small]  {$x_{i-1}$};
\draw (201.67,241.73) node [anchor=north west][inner sep=0.75pt]  [font=\small]  {$x_{i+1}$};
\draw (129,235.07) node [anchor=north west][inner sep=0.75pt]  [font=\small]  {$x_{i+4}$};
\draw (26,48.07) node [anchor=north west][inner sep=0.75pt]  [font=\small]  {$\Vert x_{i} \| \ \mathbb{B}_{2}^{d}$};
\draw (174,46.07) node [anchor=north west][inner sep=0.75pt]  [font=\small]  {$\mathbb{B}_{2}^{d}( x_{i} ,\ \delta )$};
\draw (560,38.07) node [anchor=north west][inner sep=0.75pt]  [font=\small]  {$x_{1}$};
\draw (609.67,91.07) node [anchor=north west][inner sep=0.75pt]  [font=\small]  {$x_{2}$};
\draw (415.67,6.73) node [anchor=north west][inner sep=0.75pt]  [font=\small]  {$h_{1} =h_{2}$};
\draw (594.67,196.07) node [anchor=north west][inner sep=0.75pt]  [font=\small]  {$x_{3}$};
\draw (432,214.07) node [anchor=north west][inner sep=0.75pt]  [font=\small]  {$x_{4}$};
\draw (464,69.4) node [anchor=north west][inner sep=0.75pt]  [font=\small]  {$x_{5}$};
\draw (526,130.73) node [anchor=north west][inner sep=0.75pt]  [font=\small]  {$x_{6}$};
\draw (406,268.73) node [anchor=north west][inner sep=0.75pt]  [font=\small]  {$h_{3}$};
\draw (614.67,280.73) node [anchor=north west][inner sep=0.75pt]  [font=\small]  {$h_{4}$};
\draw (345.33,151.4) node [anchor=north west][inner sep=0.75pt]  [font=\small]  {$h_{5}$};
\draw (361.33,221.4) node [anchor=north west][inner sep=0.75pt]  [font=\small]  {$h_{6}$};

\end{tikzpicture}

\caption{\textbf{Illustration of \cref{alg:det_onelayer} and \cref{thm:prob}} Given inputs $X^+, X^-$, let $x_i$ be the point with the $i$-th largest norm. The left picture illustrates how \cref{alg:det_onelayer} picks each $h_i$ according to the proof sketch. The right picture is a concrete example of this on a input.}
\label{fig:drawing}
\end{figure}

This proof is extremely technical. Using \cref{fig:drawing}, we first sketch the main idea for all three cases, before proceeding to prove each case in detail.

For each $i$, we draw two balls $\B^d_2(x_i,\delta)$ and $\|x_i\|\B_2^d$ as in the left picture in \cref{fig:drawing}. As $x_i$ is $\delta$-separated from all $x$ such that $\sigma(x)\neq \sigma(x_i)$, the points $x_j$ with higher index than $x_i$(i.e., smaller norm) and different \emph{sign} (e.g. the blue points $x_{i+1}$ and $x_{i+2}$) must lie in $\|x_i\|\B_2^d\backslash \B^d_2(x_i,\delta)$. In particular, they lie below the orange hyperplane formed by the intersection of the surfaces of the two balls. Roughly, we see that the distance between the orange hyperplane and $x_i$ is large enough compared to the choices of $\gamma$, so we can find a hyperplane $h$ through the region between $x_i$ and the orange hyperplane while maintaining distance at least $\gamma$ to both of them. We see that any such $h$ satisfies $B_i$ as all $x_j$'s lie below the orange hyperplane, so their distance to $h$ is at least that of the orange hyperplane, namely $\gamma$.

Now, we discuss how to bound probability of this occurring when $h$ is chosen randomly as in (i), (ii), and (iii). For Case (i) where $v\sim \Unif(\S^{d-1})$, we condition on $v$ being close to $x_i/\Vert x_i\Vert$. Then, $h$ will be roughly parallel to the orange hyperplane with normal vector $x_i$, so we can lower bound the probability that $t$ translates $h$ to go through the region while maintaining a $\gamma$ distance. Case (ii) follows by relating the Gaussian $v$ to the uniform $v$ in Case (i).

We also reduce Case (iii) to Case (i). To escape the curse of dimensionality, we use a trick to convert the ambient space to $\mathbb{R}^k$ with lower dimension: by \cite[Proof of Theorem 1.6]{dirksen2021separation}, $v\sim \N(0, I_d)$ is distributed identically as $G^{\intercal}u$ with independent $G\sim \N(0, I_{k\times d})$ and $u\sim \Unif(\S^{k-1})$ for all $k\in\N$. Let $A:=G/\sqrt{k}$ and $H(z):=u^{\intercal}z+t/\sqrt{k}$, so $h(z) = {H(Az)}/{\sqrt{k}}$ for all $z$. We translate the condition $B_i$ on hyperplane $h$ and points $x_i$ in $\mathbb{R}^d$ to a condition on to a similar condition on hyperplane $H$ and points $Ax_i$ in $\mathbb{R}^k$, upon which we follow the argument of Case (i). A Matrix Deviation Inequality in \cite[Exercise 9.1.8]{Ver18} is needed to show the relevant geometry is preserved passing to $H$ and $Ax_i$.

Since $\P(B_i)$ increases as $\gamma$ decreases, it suffices to show lower bounds for $\P(B_i)$ when $\gamma$ is equal to the respective upper bounds in (i), (ii), and (iii). We now prove \cref{thm:prob} rigorously for each of the three cases.

\begin{proof}[Proof of Case (i) of \cref{thm:prob}]
Here, $v\sim \Unif(\S^{d-1})$. The idea is to condition on $v$ being close to $x_i/\Vert x_i\Vert$ and find which $t$ makes $h(z)=v^{\intercal}z+t$ satisfy $B_i$. Note, for any $x_{j} \in X^{+}\cup X^{-}$ with $j >i$ and $\sigma(x_i)\neq \sigma(x_j)$, we know $\Vert x_j\Vert \leq \Vert x_{i}\Vert$ by our ordering and $ \Vert x-x_{i}\Vert\geq \delta$ by $\delta$-separation. First, for any $j$ as in the definition of $B_i$, we show $x_i^{\intercal}x_j\leq \Vert x_i\Vert ^2 - \delta^2/2$ as follows:
\[
\delta^{2} \le (x_{i}-x_{j})^{\intercal}(x_{i}-x_{j}) 
= \Vert x_{i}\Vert ^{2}+\Vert x_{j}\Vert^{2}-2x_{i}^{\intercal}x_{j}
\leq 2\Vert x_i\Vert ^{2}-2x_{i}^{\intercal}x_{j}.
\]

Now, define $\hat{x}_{i}=x_{i}/\Vert x_{i}\Vert$. For every $v\in \S^{d-1}$, we note $\hat{x}_i^{\intercal}x_i=\Vert x_i\Vert\leq R$ and bound $\left(v-\hat{x}_{i}\right)^{\intercal}x_{i}\geq-\|v-\hat{x}_{i}\|\cdot \|x_{i}\|$ by Cauchy-Schwarz to obtain
\begin{equation}\label{eq:vxi}
v^{\intercal}x_{i} = \hat{x}_{i}^{\intercal}x_{i} + \left(v-\hat{x}_{i}\right)^{\intercal}x_{i} \geq \Vert x_{i}\Vert - \Vert v-\hat{x}_{i}\Vert \cdot \Vert x_{i}\Vert\geq \Vert x_{i}\Vert - R\Vert v-\hat{x}_{i}\Vert.
\end{equation}

Now, for any $j$ as in the definition of $B_i$, write $v^{\intercal}x_{j} = \hat{x}_{i}^{\intercal}x_{j} + \left(v-\hat{x}_{i}\right)^{\intercal}x_{j}$. We had $x_i^{\intercal}x_j\leq \Vert x_i\Vert ^2 - \delta^2/2$, so $\hat{x}_i^{\intercal}x_j\leq \Vert x_i\Vert - \delta^2/2\Vert x_i\Vert$. Since $\Vert x_j\Vert \leq \Vert x_i\Vert\leq R$ and $\left(v-\hat{x}_{i}\right)^{\intercal}x_{j}\leq \Vert v-\hat{x}_{i}\Vert\cdot \Vert x_{j}\Vert$ by Cauchy-Schwarz, we obtain
\begin{equation}\label{eq:vxj}
v^{\intercal}x_{j}  \leq \Vert x_{i}\Vert -\frac{\delta^{2}}{2\Vert x_{i}\Vert} + \Vert v-\hat{x}_{i}\Vert\cdot \Vert x_{j}\Vert \leq \Vert x_{i}\Vert -\frac{\delta^{2}}{2R} + R\Vert v-\hat{x}_{i}\Vert.
\end{equation}

Recall that $h(z)=v^{\intercal}z+t$ and $B_i$ occurs if and only if $v^{\intercal}x_i+t\geq \gamma$ and $v^{\intercal}x_j+t\leq-\gamma$ for all $j$ where $j>i$ and $\sigma (x_j)\neq\sigma (x_i)$. This is equivalent to $t\geq \gamma-v^{\intercal}x_i$ and $t\leq -\gamma-v^{\intercal}x_j$ for all such $j$. We bound $\vert v^{\intercal}x_i\vert \leq \Vert v\Vert \cdot\Vert x_i\Vert \leq R$ and similarly $\vert v^{\intercal}x_{j}\vert \leq R$. Hence, for a fixed $v$, $B_i$ occurs if
\[ t\in I:=\left[\gamma - v^{\intercal}x_i, \  -\gamma - \max_{j:j>i, \sigma(x_j)\neq\sigma(x_i)}v^{\intercal}x_j\right] \subset [-R, R]\subset [-\lambda, \lambda] \]
as $\lambda\geq R$. Hence, we bound $\P(B_i\mid v)$ below from by the length of $I$ over $2\lambda$, i.e.
\[ \P_{t}\left(B_{i}\mid v\right) \geq \frac{1}{2\lambda}\left(v^{\intercal}x_{i}-\max_{j:j>i, \sigma(x_j)\neq \sigma(x_i)}( v^{\intercal}x_{j}) -2\gamma\right)\geq \frac{1}{2\lambda}\left(\frac{\delta^{2}}{2 R}-2R\Vert v-\hat{x}_{i}\Vert -2\gamma\right)\]
where the second inequality follows from \eqref{eq:vxi} and \eqref{eq:vxj}. Let $r:=\left({\delta^{2}}/{4R^{2}}\right)\left(1-{1}/{d}\right)$. Now, $r\in [0,2]$ as $\delta \leq 2R$. As $v\sim \Unif(\S^{d-1})$, applying \cref{lem:geo lm}(1) gives
\[
\P(B_{i}) \geq \P\left(B_{i}\mid \Vert v-\hat{x}_{i}\Vert \leq r\right)\cdot \P_{t}\left(\Vert v-\hat{x}_{i}\Vert \leq r\right)
\geq \frac{1}{2\lambda}\left(\frac{\delta^{2}}{2R}-2Rr-2\gamma\right)\left(\frac{1}{2}\right)\left(\frac{r}{2}\right)^{d-1}.
\]
Plug in $r$ and bound $\gamma \leq \delta^{2}/8Rd$ to obtain
\begin{align*}
\P(B_{i}) & \geq \frac{1}{4\lambda}\left[\frac{\delta^{2}}{2R}-2R\cdot \frac{\delta^{2}}{4R^{2}}\left(1-\frac{1}{d}\right)-2\cdot \frac{\delta^{2}}{8Rd} \right]\left[\frac{\delta^{2}}{8R^{2}}\left(1-\frac{1}{d}\right)\right]^{d-1}
\\ &  = \frac{1}{4\lambda}\left(\frac{\delta^{2}}{4Rd}\right)\left(\frac{\delta^{2}}{8R^{2}}\right)^{d-1}\left(1-\frac{1}{d}\right)^{d-1} 
\\ & \geq \frac{R}{8(d-1)\lambda}\left(\frac{\delta^{2}}{8R^{2}}\right)^{d}
\\ & = p
\end{align*}
where the second inequality follows from the fact that $(1-1/d)^d$ is increasing in $d$ for $d\geq 2$, so it is at least its value $1/4$ at $d=2$. This establishes Case (i).
\end{proof} 
\begin{proof}[Proof of Case (ii) of \cref{thm:prob}]
Here, $v\sim \N(0, I_{d})$. By \cref{lem:prob lm}(1), we can change variables to $\rho = \Vert v\Vert$ with $\rho^2\sim \chi^2_d$ and $\hat{v}=v/\rho\sim \Unif(\S^{d-1})$, which are independent. For any $1\leq \rho\leq 3\sqrt{d}$, we compute
\begin{align*}
\P(B_i\mid \rho) & = \P\left(\hat{v}^{\intercal}x+t/\rho \geq \gamma/\rho \ \forall \ x\in X^+ \text{ and } \hat{v}^{\intercal}x+t/\rho \leq -\gamma/\rho \ \forall \ x\in X^-\right)
\\ & \geq \P\left(\hat{v}^{\intercal}x+t/\rho \geq \gamma \ \forall \ x\in X^+ \text{ and } \hat{v}^{\intercal}x+t/\rho \leq -\gamma \ \forall \ x\in X^-\right)
\\ & \geq \P\left(\hat{v}^{\intercal}x+t/\rho\text{ satisfies event }B_i\right).
\end{align*}
Now, $\hat{v}\sim\Unif(\S^{d-1})$ and, for a fixed $\rho$, $t/\rho \sim \Unif([-\lambda/\rho, \lambda/\rho]$. Since $\rho\leq 3\sqrt{d}$ and $\lambda\geq 3\sqrt{d}R$, $\lambda/\rho\geq R$. Hence, Case (i) applies to show 
\[ \P(B_i\mid \rho) \geq \P\left(\hat{v}^{\intercal}x+t/\rho\text{ satisfies event }B_i\right) \geq p.\]
This holds for every $\rho$ where $1\leq\rho\leq 3\sqrt{d}$. By conditioning on it, we obtain

\begin{align*}
	\P(B_i) & \geq \P\left(B_i\mid 1\leq \rho\leq 3\sqrt{d}\right)\cdot \P\left(1\leq \rho\leq 3\sqrt{d}\right)
	\\ & \geq \frac{1}{10}\inf\left\lbrace \P(B_i\mid\rho) : {1\leq \rho\leq 3\sqrt{d}}\right\rbrace 
	\\ & \geq \frac{p}{10}
\end{align*}
where $ \P(1\leq \rho\leq 3\sqrt{d})\geq 1/10$ by \cref{prop:chi}, showing Case (ii).
\end{proof}
\begin{proof}[Proof of Case (iii) of \cref{thm:prob}]
Following the proof sketch, we will first choose $k$ using \cite[Exercise 9.1.8]{Ver18}, so the event $E$ where the $\delta$-separation and ordering of $x_i$ is preserved happens with nontrivial probability. Then, we translate condition $B_i$ to ambient space $\mathbb{R}^k$, and show $\mathbb{P}(B_i\mid E)$ is nontrivial. If $\|x_i\|$ is large, we will do a similar analysis as Case (i) in $\R^k$; if $\|x_i\|$ is small, we will calculate $\P(B_i)$ directly and see it is much larger than when $\|x_i\|$ is large.

\paragraph{Step 1: Choosing Dimension $k$}
For every positive integer $k$, $v\sim \N(0, I_d)$ is distributed identically as $G^{\intercal}u$ with independent $G\sim \N(0, I_{k\times d})$ and $u\sim \Unif(\S^{k-1})$. We can require $C$ to be such that $k$ defined in \eqref{eq:k} satisfy $k\in\Z_{\geq 2}$.

To specify $k$, let $A:=G/\sqrt{k}$. By \cite[Exercise 9.1.8]{Ver18}, for any constants $\theta >0$ and $\omega\in [0,1)$, there exists $C$ in the definition of $k$ in \eqref{eq:k} such that with probability at least $\omega$, for all $x\in X^+\cup X^-\cup (X^+-X^-)$,
\begin{equation}\label{eq:E}
    \vert \Vert Ax\Vert - \Vert x\Vert \vert \leq \theta.
\end{equation}
Fix $\theta:= {\delta^2}/{32R}$ and $\omega:= 8/9$, and denote the event of the above display as $E$. We can choose $C$ in \eqref{eq:k} such that $k\in\Z_{\geq 2}$ and $\P(E)\geq 8/9$. Note, $C$ depends on $\omega$ and $\theta$, which depends on $R$ and $\delta$.

\paragraph{Step 2: Relating to Case (i)}
Let $H(z):=u^{\intercal}z+t/\sqrt{k}$. For every $1\leq j\leq N$, let $y_j:=Ax_j\in\R^k$ and $\sigma (y_j):=\sigma (x_j)$. Note that for all $z$, $h(z) = \sqrt{k} H(Az)$, so $B_i$ holds if and only if $H(y_i)\geq \gamma/\sqrt{k}$ and $H(y_j)\leq -\gamma/\sqrt{k}$ for all $j>i$ and $\sigma(y_i)\neq \sigma(y_j)$. Let $B_i'$ be $B_i$ conditioned on $E$. Note
\[\P(B_i)\geq \P(B_i\mid E)\cdot \P(E) \geq \frac{8}{9} \cdot \P(B_i\mid E).\]

For the rest of the proof, we lower bound $\P(B_i\mid E)$, i.e. the probability that $H(y_i)\geq \gamma/\sqrt{k}$ and $H(y_j)\leq -\gamma/\sqrt{k}$ for all $j>i$ and $\sigma(y_i)\neq \sigma(y_j)$ given $E$. By the ordering of the points by their norms, we first notice that for all $j>i$, 
\begin{equation}\label{eq:norms}
    \Vert x_i\Vert +\theta \geq \Vert y_i\Vert = \Vert Ax_i\Vert \geq  \Vert x_i\Vert  -\theta \geq  \Vert x_j\Vert-\theta \geq \Vert Ax_j\Vert -2\theta =\Vert y_j\Vert-2\theta.
\end{equation}

On the other hand,  since $AX^+$ and $AX^-$ are $(\delta -\theta)$-separable, so $\Vert y_i -y_j\Vert\geq \delta -\theta$ whenever $\sigma (y_i)\neq\sigma (y_j)$. These conditions are similar to the $\delta$-separation assumption and ordering of $x_i$ in Case (i). We show Case (iii) in two sub-cases Case (iiia) and Case (iiib) according to whether $\Vert x_i\Vert \geq\delta/2$ or $\|x_i\|\leq \delta/2$.

\paragraph{Step 3: Case (iiia)}
In this case, assume $\ell:=\Vert x_i\Vert\geq \delta/2$. We follow Case (i) applied to the $y_i$'s. On the event $E$ defined in \eqref{eq:E}, for every $y_j$ with $j>i$ and $\sigma(y_j)\neq \sigma (y_i)$, by \eqref{eq:norms} and $(\delta-\theta)$-separation
\[
(\delta-\theta)^{2}\leq (y_{i}-y_{j})^{\intercal}(y_{i}-y_{j}) = \Vert y_{i}\Vert ^{2}+\Vert y_{j}\Vert^{2}-2y_{i}^{\intercal}y_{j} \leq 2(\ell +\theta) ^{2}-2y_{i}^{\intercal}y_{j}
\]
so we have $y_{i}^{\intercal}y_{j}\leq (\ell+\theta)^2-(\delta - \theta)^{2}/2$. Let $\hat{y}_{i}=y_{i}/\Vert y_{i}\Vert$. For every $u\in \S^{k-1}$, we bound $(u-\hat{y}_i)^{\intercal}y_i\geq -\Vert u-\hat{y}_i\Vert \cdot \Vert y_i\Vert $ by Cauchy-Schwarz and $\ell-\theta\leq \Vert y_i\Vert \leq \ell+\theta$ from \eqref{eq:norms} to obtain
\begin{equation}\label{eq:uyi}
  u^{\intercal}y_{i} = \hat{y}_{i}^{\intercal}y_{i} + \left(u-\hat{y}_{i}\right)^{\intercal}y_{i} \geq \Vert y_{i}\Vert - \Vert u-\hat{y}_{i}\Vert \cdot \Vert y_{i}\Vert\geq (\ell-\theta) - (\ell+\theta)\Vert u-\hat{y}_{i}\Vert   
\end{equation}
Similarly, for any $j$ where $j>i$ and $\sigma (y_j)\neq\sigma(y_i)$, we had $y_{i}^{\intercal}y_{j}\leq (\ell+\theta)^2-(\delta - \theta)^{2}/2$, so $\hat{y}_{i}^{\intercal}y_{j}\leq (\ell+\theta)^2/\Vert y_i\Vert -(\delta - \theta)^{2}/2\Vert y_i\Vert$. Combined with the bound $(u-\hat{y}_i)^{\intercal}y_j\leq \Vert u-\hat{y}_i\Vert \cdot \Vert y_j\Vert $ by Cauchy-Schwarz inequality, we obtain the former inequality in the following
\begin{align}\label{eq:uyj}
u^{\intercal}y_{j} & = \hat{y}_{i}^{\intercal}y_{j} + \left(u-\hat{y}_{i}\right)^{\intercal}y_{j} \nonumber
\\ & \leq \frac{(\ell+\theta)^2}{\Vert y_i\Vert } - \frac{(\delta-\theta)^2}{2\Vert y_i\Vert }+ \Vert u-\hat{y}_{i}\Vert\cdot \Vert y_{j}\Vert \nonumber
\\ & \leq \frac{(\ell+\theta)^2}{\ell-\theta} - \frac{(\delta -\theta)^2}{2(\ell+\theta)}+(\ell+\theta)\Vert u-\hat{y}_i\Vert
\end{align}
where we use $\ell-\theta\leq \Vert y_i\Vert\leq \ell+\theta $ and $\Vert y_j\Vert \leq \ell+\theta$ from \eqref{eq:norms} for the latter.

Recall that $H(z)=u^{\intercal}z+t/\sqrt{k}$ and $B_i$ occurs if and only if $u^{\intercal}x_i+t/\sqrt{k}\geq \gamma/\sqrt{k}$ and $u^{\intercal}x_j+t/\sqrt{k}\leq-\gamma/\sqrt{k}$ for all $j$ where $j>i$ and $\sigma (y_j)\neq\sigma (y_i)$. This is equivalent to $t\geq \gamma-\sqrt{k}u^{\intercal}y_i$ and $t\leq -\gamma-\sqrt{k}u^{\intercal}y_j$ for all such $j$. Then, for a fixed $u$, the event $B_i$ occurs if
\[ t\in I : = \left[\gamma - \sqrt{k}u^{\intercal}y_i, \  -\gamma -\sqrt{k}\max_{j:j>i, \sigma(y_j)\neq\sigma(y_i)}u^{\intercal}y_j\right]. \]
We bound $\vert u^{\intercal}y_i\vert \leq \Vert u\Vert \cdot\Vert y_i\Vert \leq  \Vert x_i\Vert +\theta \leq R+\theta$ and similarly $\vert u^{\intercal}y_{j}\vert \leq R+\theta$.
Since $\theta = \delta^2/32R\leq (2R)^2/32R= R/8$ and $\lambda\geq 9R\sqrt{k}/8$, we have
\[ I\subset [-\sqrt{k}(R+\theta), \sqrt{k}(R+\theta)]\subset [-9\sqrt{k}R/8, 9\sqrt{k}R/8]\subset [-\lambda, \lambda].\]
So we may now bound $\P(B_i\mid E, u)$ from below by the length of $I$ over $2\lambda$, i.e.
\begin{align*}
\P_{t}\left(B_{i}\mid E, u\right) & \geq \frac{\sqrt{k}}{2\lambda}\left(u^{\intercal}y_{i}-\max_{j:j>i, \sigma(y_j)\neq \sigma(y_i)}( u^{\intercal}y_{j}) -\frac{2\gamma}{\sqrt{k}}\right)
\\ & \geq \frac{\sqrt{k}}{2\lambda}\left((\ell-\theta) -\frac{(\ell+\theta)^2}{\ell-\theta} + \frac{(\delta -\theta)^2}{2(\ell+\theta)} - 2 (\ell+\theta)\Vert u-\hat{y}_{i}\Vert - \frac{2\gamma}{\sqrt{k}}\right)  
\\ & = \frac{\sqrt{k}}{\lambda}\left(\frac{(\delta -\theta)^2}{4(\ell+\theta)} -\frac{2\ell\theta}{\ell-\theta}  - (\ell+\theta)\Vert u-\hat{y}_{i}\Vert - \frac{\gamma}{\sqrt{k}}\right) 
\\ & = \frac{\sqrt{k}}{\lambda}\left(\frac{(\delta -\theta)^2}{4(\ell+\theta)} -\frac{2 (\delta^2/32R)}{1-\theta/\ell}  - (\ell+\theta)\Vert u-\hat{y}_{i}\Vert - \frac{\gamma}{\sqrt{k}}\right) 
\end{align*}
where we plug in \eqref{eq:uyi} and \eqref{eq:uyj} for the second inequality and the last equality is due to $\theta = \delta^2/32R$. We know $\delta /2 \leq \ell \leq R$, so we have  $\theta \leq \delta/16 $. Note that the expression inside the parenthesis of the last line of the above display  decreases when we replace every $\theta$ by its upper bound $\delta/16$, replace $\ell$ in the first and third terms by its upper bound $R$, and replace $\ell$ in the second term by its lower bound $\delta/2$. By these substitutions, we obtain the following inequality
\begin{align*}
\P(B_{i}\mid E, u) & \geq \frac{\sqrt{k}}{\lambda}\left(\frac{(15\delta/16)^2}{4(R+\delta/16)} -\frac{2 (\delta^2/32R)}{1-(\delta/16)/(\delta/2)}  - (R+\delta/16)\Vert u-\hat{y}_{i}\Vert - \frac{\gamma}{\sqrt{k}}\right)
\\ & \geq \frac{\sqrt{k}}{\lambda}\left(\frac{(15\delta/16)^2}{4(R+R/8)} -\frac{2(\delta^2/32R)}{1-(\delta/16)/(\delta/2)}  - (R+R/8)\Vert u-\hat{y}_{i}\Vert - \frac{\gamma}{\sqrt{k}}\right) 
\\ & = \frac{\sqrt{k}}{\lambda}\left(\frac{111}{896}\cdot\frac{\delta^2}{R} - (9R/8)\cdot\Vert u-\hat{y}_{i}\Vert - \frac{\gamma}{\sqrt{k}}\right) 
\\ & \geq \frac{\sqrt{k}}{\lambda}\left(\frac{\delta^2}{18R} - (9R/8)\cdot\Vert u-\hat{y}_{i}\Vert\right) 
\end{align*}
where the second inequality follows by replacing $\delta$ by its upper bound $2R$, and the last inequality follows $\gamma \leq \delta^2\sqrt{k}/18R$ by bounding $111/896\geq 1/9$.

We condition on $\Vert u-y_i\Vert \leq r:=(4\delta^{2}/81R^{2})(1-1/k)\in [0, 2]$ to obtain the first inequality below. As $u\sim \Unif(\S^{k-1})$, \cref{lem:geo lm}(1) implies the second inequality below, i.e.
\begin{align*}
\P(B_{i}\mid E) & \geq\P\left(B_{i}\mid E, \Vert u-\hat{y}_{i}\Vert \leq r\right)\cdot \P_{t}\left(\Vert u-\hat{y}_{i}\Vert \leq r\right)
\\ & \geq \frac{\sqrt{k}}{\lambda}\left(\frac{\delta^2}{18R} - \frac{9Rr}{8}\right) \cdot \left(\frac{1}{2}\right)\left(\frac{r}{2}\right)^{k-1}
\\ & = \frac{\sqrt{k}}{\lambda}\left(\frac{\delta^2}{18R} - \frac{9R}{8}\cdot \frac{4\delta^2}{81R^2}\left(1-\frac{1}{k}\right)\right) \left(\frac{1}{2}\right)\left(\frac{2\delta^2}{81R^2}\left(1-\frac{1}{k}\right)\right)^{k-1}
\\ & = \frac{\sqrt{k}}{2\lambda}\left(\frac{\delta^2}{18Rk}\right) \left(\frac{2\delta^2}{81R^2}\right)^{k-1}\left(1-\frac{1}{k}\right)^{k-1}
\\ & \geq \frac{\sqrt{k}}{2\lambda}\left(\frac{9R}{4k}\right) \left(\frac{2\delta^2}{81R^2}\right)^{k}\left(1-\frac{1}{k}\right)^{-1}\left(\frac{1}{4}\right)
\\ & \geq  \frac{9}{32}\left(\frac{R}{\lambda\sqrt{k}}\right) \left(\frac{2\delta^2}{81R^2}\right)^{k}
\end{align*}
where the third inequality follows from the fact that $(1-1/k)^k$ is increasing in $k$ for $k\geq 2$, so it can be lower bounded by its value $1/4$ at $k=2$, and the last inequality is due to $(1-1/k)^{-1}\geq 1$. Hence
\[\P(B_i)\geq \P(E)\cdot \P(B_i\mid E) \geq \frac{8}{9}\cdot \frac{9}{32}\left(\frac{R}{\lambda\sqrt{k}}\right) \left(\frac{2\delta^2}{81R^2}\right)^{k} = \frac{R}{4\lambda\sqrt{k}} \left(\frac{2\delta^2}{81R^2}\right)^{k} =:q \]
establishing Case (iiia), as desired.

\paragraph{Step 4: Case (iiib)}
In this case, we have  $\ell := \Vert x_i\Vert \leq\delta/2$. By the triangle inequality, $\B_2^d(x_i, \delta)\supset (\delta/2) \B_2^d\supset \Vert x_i\Vert \B_2^d$, so there cannot be any $x_j$ with smaller norm in the other set as $x_i$. Then, $B_i = \lbrace h(x_i)\geq \gamma\rbrace$, which intuitively happens with much higher probability than $q$. Continuing with the same notation as in Case (iiia), we have that conditioned on the event $E$ (recall from \eqref{eq:E})
\[B_i\iff h(x_i)\geq \gamma \iff H(y_i) = u^{\intercal}y_i+t/\sqrt{k}\geq \gamma/\sqrt{k}\iff t\geq \gamma -\sqrt{k}u^{\intercal}y_i.\]
As $u\sim\Unif(\S^{k-1})$, for any $y_i \in \mathbb{R}^k$, we see $\P(u^{\intercal}y_i\geq 0)=1/2$. Now, we get
\[
    \P(B_i\mid E) \geq \P\left(t\geq \gamma -\sqrt{k}u^{\intercal}y_i\mid E, u^{\intercal}y_i\geq 0\right)\cdot \P( u^{\intercal}y_i\geq 0) \geq \frac{1}{2} \cdot \P\left(t\geq \gamma\mid E\right)
\]
Since the event $E$ is independent of the random variable $t$, $\P\left(t\geq \gamma\mid E\right) =\P\left(t\geq \gamma\right)$. Since $t\sim\Unif([-\lambda, \lambda])$, $\lambda\geq 9R\sqrt{k}/8$, and $\gamma \leq \delta^2/18R\sqrt{k}$, we obtain the first inequality below
\[ \P(t\geq \gamma)= \frac{\lambda-\gamma}{2\lambda}\geq \frac{1}{2}-\frac{\delta^2/18R\sqrt{k}}{2(9R\sqrt{k})/8}=\frac{1}{2}-\frac{2\delta^2}{81kR^2}\geq \frac{1}{2}-\frac{2(2R)^2}{162R^2}> \frac{1}{3}\]
where the second inequality follows from $k\geq 2$ and $\delta \leq 2R$. Plugging this back into the equation above gives $\P(B_i\mid E)\geq 1/6$. Finally, condition on $E$ to get
\[ \P(B_i)\geq \P(B_i\mid E)\cdot \P(E)\geq \frac{1}{6}\cdot \frac{8}{9} = \frac{4}{27}\]

On the other hand, using $\lambda \geq 9R\sqrt{k}/8\geq R\sqrt{k}$ and $\delta \leq 2R$, we obtain the first inequality below
\[
q = \frac{R}{4\lambda\sqrt{k}}\left(\frac{2\delta^2}{81R^2}\right)^k \leq \frac{R}{4Rk}\left(\frac{2(2R)^2}{81R^2}\right)^2 \leq \frac{1}{8}\left(\frac{8}{81}\right)^2\leq \frac{4}{27}\leq \P(B_i)
\]
where the second inequality follows since $k\geq 2$. This proves Case (iiib).
\end{proof}
\subsection{Showing $\Phi$ Makes Finite Sets Linearly Separable}
Assuming the existence of hyperplanes $h_i$ as proven in \cref{thm:prob}, we now argue the correctness of the algorithm and bound the margin of separation, thereby showing that one deterministic layer separates finite sets.

\begin{theorem}\label{thm:fin}
For output $\Phi$ of \cref{alg:det_onelayer} on any sets $X^{+}, X^{-}$ and parameter $\gamma$ satisfying the input conditions, $\Phi (X^{+})$ and $\Phi (X^{-})$ are linearly separable with margin at least $\sqrt{N}\mathcal{M}(\gamma, N)$ in \eqref{eq:p}.
\end{theorem}

Define $x_i$ and $h_j$ as in \cref{alg:det_onelayer} (exists by \cref{thm:prob}) and $\sigma$ as in \eqref{eq:sigma}. We state an algorithm that finds a hyperplane separating $\Phi(X^+)$ and $\Phi(X^-)$ with the desired margin, before proving these guarantees.
\begin{algorithm}[H]
\caption{Separate the image of the sets under the output of \cref{alg:det_onelayer}}
\label{alg:weights}
\begin{algorithmic}[1]
\Require Two $\delta$-separated finite sets $X^{+}, X^{-}\subset R\B_2^d$ and parameter $\gamma$ satisfying the input conditions of \cref{alg:det_onelayer}, and the output $\Phi$ of \cref{alg:det_onelayer} on $X^{+}, X^{-}$, and $\gamma$.
\Ensure A hyperplane $H\subset \R^N$ separating $\Phi(X^+)$ and $\Phi(X^-)$ with margin at least $\sqrt{N}\mathcal{M}(\gamma, N)$.
\For{$i = N,N-1,\ldots,1$}
    \State $a_i\gets\sigma(x_i)\left(1+\left\vert\sum_{j=i+1}^Na_j(\Phi(x_i))_j\right\vert\right)/\gamma$.
\EndFor
\State \Return the hyperplane $H(z)=a^{\intercal}z$ where $a=(a_i)_i\in\R^N$.
\end{algorithmic}
\end{algorithm}
\begin{proof}
By definition of $h_j$ in \cref{alg:det_onelayer} and event $B_j$, we have that 
\begin{equation}\label{eq:hj}
h_{j}(x_{i})=\begin{cases}
\leq -\gamma & \textup{if $j<i$ and $\sigma(x_j) \neq \sigma(x_i) $}\\
\geq \gamma  & \textup{if $j=i$}\\
* & \textup{otherwise}
\end{cases}
\end{equation}
where $*$ denotes unknown reals. Hence, after activation,
\begin{equation}\label{eq:phij}
(\Phi(x_i))_j=\Re(h_{j}(x_{i}))=\begin{cases}
0  & \textup{if $j<i$ and $\sigma(x_j) \neq \sigma(x_i) $}\\
\geq \gamma  & \textup{if $j=i$}\\
\geq 0 & \textup{otherwise}
\end{cases}.
\end{equation}

As in \cref{alg:weights}, we choose the weight vector $a\in\R^N$ by picking $a_{i}$ recursively for $i=N, N-1, \dots , 1$ as follows:
\begin{equation}\label{eq:weight_vec_choice}
    a_{i}=\frac{\sigma(x_i)}{\gamma }\left(1+\left\vert\sum_{j=i+1}^{N}a_{j}(\Phi(x_{i}))_{j}\right\vert\right).
\end{equation}
Note that $a_i$ has same sign as $\sigma(x_i)$. Let $H(z) = a^{\intercal}z$ be an affine function associated with the hyperplane $\{z : H(z) = 0\}$. We first claim and prove that $H$ separates the points of $\Phi(X^{+})$ and $\Phi(X^{-})$, and then derive the lower bound on the margin of separation.

\begin{claim}\label{claim:H sep}
$H$ separates $\Phi(X^{+})$ and $\Phi(X^{-})$ with margin at least $1/\Vert a\Vert$.
\end{claim}
\begin{proof}[Proof of \cref{claim:H sep}]
It suffices to show $H (\Phi(x_{i}))\geq 1$ if $x_{i}\in X^{+}$ and $H (\Phi(x_{i}))\leq -1$  if $x_{i}\in X^{-}$. For every $1 \le i \le N$, we have
\begin{align}\label{eq:4terms}
\sigma(x_i)  H(\Phi(x_{i})) & = \sigma(x_i) \sum_{j=1}^{N}a_{j}(\Phi(x_{i}))_{j} \nonumber
\\ & = \sigma(x_i) \sum_{j<i, \sigma(x_j) =\sigma(x_i) }a_{j}(\Phi(x_{i}))_{j} + \sigma(x_i) \sum_{j<i, \sigma(x_j) \neq \sigma(x_i) }a_{j}(\Phi(x_{i}))_{j}\nonumber
\\ & \qquad + \sigma(x_i) a_{i}(\Phi(x_{i}))_{i} + \sigma(x_i) \sum_{j>i}a_{j}(\Phi(x_{i}))_{j}
\end{align}
where the second equality is obtained by splitting the summation into four groups depending on whether $j < i$, $j = i,$ or $j > i$, and depending on whether $\sigma(x_j) = \sigma(x_i)$. We consider each quantity.

For each term in the first quantity, $a_j$ has the same sign as $\sigma(x_j)=\sigma(x_i)$, so
\begin{equation}\label{eq:first_term}
    \sigma(x_i) \sum_{j<i, \sigma(x_j) =\sigma(x_i) }a_{j}(\Phi(x_{i}))_{j} = \sum_{j<i, \sigma(x_j) =\sigma(x_i) }\sigma(x_j)a_{j}(\Phi(x_{i}))_{j} \ge 0.
\end{equation}
For each term in the first quantity, $(\Phi(x_i))_j = 0$ if $j<i$ and $\sigma(x_j) \neq \sigma(x_i)$, so
\begin{equation}\label{eq:second_term}
    \sigma(x_i) \sum_{j<i, \sigma(x_j) \neq \sigma(x_i) }a_{j}(\Phi(x_{i}))_{j} = 0.
\end{equation}
For the third quantity, from \eqref{eq:phij} we have $(\Phi(x_{i}))_{i} \ge \gamma$, and from \eqref{eq:weight_vec_choice}, so
\begin{equation}\label{eq:third_term}
    \sigma(x_i) a_{i}(\Phi(x_{i}))_{i} \ge \frac{\sigma(x_i) ^{2}}{\gamma}\left(1+\left|\sum_{j>i}a_{j}(\Phi(x_{i}))_{j}\right|\right)\gamma = 1+\left|\sum_{j>i}a_{j}(\Phi(x_{i}))_{j}\right|.
\end{equation}
Plugging \eqref{eq:first_term}, \eqref{eq:second_term}, and \eqref{eq:third_term} into \eqref{eq:4terms} gives the first inequality below
\[
\sigma(x_i)  H(\Phi(x_{i}))  \geq 0 + 0 + 1+\left|\sum_{j>i}a_{j}(\Phi(x_{i}))_{j}\right| + \sigma(x_i) \sum_{j>i}a_{j}(\Phi(x_{i}))_{j}
 \geq 1,
\]
and the second inequality follows by the triangle inequality. This shows $H (\Phi(x_{i}))\geq 1$ if $x_{i}\in X^{+}$ and $H (\Phi(x_{i}))\leq -1$  if $x_{i}\in X^{-}$, so the margin is at least $1/\Vert a\Vert$.
\end{proof}

We now further lower bound the margin of separation $1/\Vert a\Vert$.

\begin{claim}\label{claim:margin bd}
For $a$ defined in \eqref{eq:weight_vec_choice} and $\mathcal{M}$ defined in \eqref{eq:p}, $1/\Vert a\Vert \geq \sqrt{N}\mathcal{M}(\gamma, N)$.
\end{claim}
\begin{proof}[Proof of \cref{claim:margin bd}]
We first show that $(\Phi(x_i))_{j}\leq 2R$ for all $j>i$ such that $\sigma(x_i)\neq \sigma(x_j)$. We start by showing  for all $i$, $h_{i}(x)=0$ for some $x\in R\B_{2}^{d}$. This just follows from the observations that each of our hyperplanes $\lbrace h_i=0\rbrace$ intersects $R\B_2^d$ by assumption. More rigorously, consider the two cases on line 5 of \cref{alg:det_onelayer}, either way we have $h_{i}(x_{i})>0$ and $h_{i}(x_{j})\leq 0$ for some $x_{i}, x_{j}\in R\B_{2}^{d}$. Then, by the intermediate value theorem, there is some $x$ on the segment $x_{i}x_{j}$ such that $h_{i}(x)=0$. Since $R\B_{2}^{d}$ is convex, $x\in R\B^{d}_{2}$. Then, for every $y\in R\B_{2}^{d}$, since $w_{i}\in \S^{d-1}$ we have
\begin{equation}\label{eq:hiy}
\vert h_{i}(y)\vert \leq \vert h_{i}(x)\vert + \vert h_{i}(y)-h_{i}(x)\vert 
 \leq \vert w_{i}^{\intercal}(y-x)\vert 
\leq \Vert w_{i}\Vert \cdot \Vert y-x\Vert
\le 2R.
\end{equation}
Here, we used the triangle inequality to obtain the first inequality, used that $h_i(x) = 0$ to get the second inequality, applied the Cauchy-Schwarz Inequality in the third inequality, and used the fact that $x,y \in R\B_2^d$ to get the final inequality. By using the upper bound in \eqref{eq:hiy}, 
\[
\vert a_{i}\vert = \frac{1}{\gamma }\left( 1+\left|\sum_{j=i+1}^{N}a_{j}(\Phi(x_{i}))_{j}\right|\right)
\leq \frac{1}{\gamma} \left(1+2R \sum_{j=i+1}^{N}\vert a_{j}\vert\right),
\]
where we used the triangle inequality to obtain the last inequality. Thus, $\vert a_{i}\vert \leq \tilde{a}_{i}$ where $\tilde{a}_{i}$ is a sequence defined recursively by $\tilde{a}_{N}= 1/\gamma = \vert a_{N}\vert$ and 
\[
    \tilde{a}_{i} = \frac{1}{\gamma}\left(1 + 2R \sum_{j=i+1}^N \tilde{a}_j\right).
\]
We compute $\tilde{a}_{i}$ explicitly. Taking the successive differences
\[
    \tilde{a}_{i-1}-\tilde{a}_{i} = \frac{1}{\gamma}\left(1 + 2R \sum_{j=i}^N \tilde{a}_j\right) - \frac{1}{\gamma}\left(1 + 2R \sum_{j=i+1}^N \tilde{a}_j\right) = \frac{2R\tilde{a}_{i}}{\gamma},
\]
 and using the fact that $\tilde{a}_{N}= 1/\gamma$, we get
\[
    \tilde{a}_{i} = \frac{1}{\gamma}\left(1+\frac{2R}{\gamma}\right)^{N-i}.
\]
Hence,
\[
\Vert a\Vert ^{2} \leq \sum_{i=1}^{N}\frac{1}{\gamma^{2}}\left(1+\frac{2R}{\gamma}\right)^{2(N-i)} = \frac{1}{\gamma^{2}} \cdot \frac{(1+2R/\gamma)^{2N}-1}{(1+2R/\gamma )^{2}-1} = \frac{(1+2R/\gamma)^{2N}-1}{4R(R+\gamma)},
\]
which gives the desired bound by rearranging.
\end{proof}
Therefore, \cref{claim:H sep} and \cref{claim:margin bd} together implies that $H$ separates $\Phi(X^+)$ and $\Phi(X^-)$ by a margin at least $\sqrt{N}\mathcal{M}(\gamma, N)$, as desired.
\end{proof}
\section{Generalizations to Arbitrary Sets and Random Initialization}\label{sec:Two}
\subsection{Passing to Arbitrary $\delta$-separated Sets}
We first generalize \cref{thm:fin} to arbitrary sets $X^+$ and $X^-$ which may not be finite. With carefully chosen margin parameter $\gamma$ and mutual complexity parameters, if the output of \cref{alg:det_onelayer} separates the centers in \cref{def:complexity}, the next theorem shows that it must separate the entire sets $X^+$ and $X^-$.
\begin{theorem}\label{thm:arb}
Let $X^{+}, X^{-}\subset R\B_{2}^{d}$ be any two $\delta$-separated sets and $k$ as defined in \eqref{eq:k}. For any parameter $\gamma\in [0, \max(\delta^2/8Rd, \delta^2/18R\sqrt{k})]$, let $N$ be the $(\delta, 8R^2/\gamma )$-mutual complexity of $X^+$ and $X^-$. Let centers $C^{+}$ and $C^{-}$ be as guaranteed by \cref{def:complexity}. Recall $\mathcal{M}$ from \eqref{eq:p}. We have
\begin{enumerate}
    \item[(i)] $C^+$ and $C^-$ with parameter $\gamma$ satisfy the input conditions of \cref{alg:det_onelayer}.
    \item[(ii)] Let $\Phi$ be the output of \cref{alg:det_onelayer} on $C^+$ and $C^-$ with parameter $\gamma$. Then, $\Phi(X^+)$ and $\Phi(X^-)$ are linearly separable with margin at least $\sqrt{N}\mathcal{M}(\gamma/2, N)$.
\end{enumerate}
\end{theorem}
Let us briefly outline the proof and put it in context of our previous results.
(i) follows from \cref{def:complexity} and the assumptions on $X^+, X^-$, and $\gamma$. For (ii), the radii used in \cref{def:complexity} for $(\delta, 8R^2/\gamma )$-mutual complexity can be bounded above by $\gamma/2$. By \cref{thm:fin} on the finite sets $C^+$ and $C^-$, for each center of $C^{+}$, we get a hyperplane that is $\gamma$ away from that center and any center from $C^{-}$ with smaller norm, and vice versa. Then, those hyperplanes are at least $\gamma/2$ away from any point in the balls guaranteed by \cref{def:complexity} around those centers. The result now follows from the proof of \cref{thm:fin}.

\begin{proof}
	Applying \cref{def:complexity} to that sets $X^+$ and $X^-$ have $(\delta, 8R^2/\gamma)$-mutual complexity $N$. Since $C^+, C^-\subset R\B_2^d$ are $\delta$-separated, for any center $c_i^\pm$
\[ \delta\leq \dist \left(c_{i}^\pm, C^{\mp}\right)\leq 2R.\] 
Then, for any radius $r_i^\pm$, we have that
\begin{equation}\label{eq:ri}
r_i^\pm \leq \frac{\gamma}{8R^2}\dist \left(c_{i}^\pm, C^{\mp}\right)^2 \leq \frac{\gamma}{8R^2} (2R)^2=\frac{\gamma}{2}.
\end{equation}
We re-order the points in $C^+\cup C^-$ as in \cref{alg:det_onelayer}, i.e., for every $i$, let $c_i$ be the point in $C^+\cup C^-$ with the $i$-th largest norm. Similarly, let $r_i$ now be the radius associated to $c_i$. Hence, \eqref{eq:ri} implies $r_i\leq \gamma/2$ for all $i$. Also, let $\sigma (c_i)=+1$ if $c_i\in C^+$ and $\sigma (c_i)=-1$ if $c_i\in C^-$ as in \eqref{eq:sigma}. Therefore, the first condition of \cref{def:complexity} says that $c_i\in \conv(X_i)$ where $X_i:=X^{\sigma(c_i)}\cap \B_2^d(c_i, r_i)$.

\paragraph{Proof of Part (i)}
Here we show that if the input conditions of \cref{alg:det_onelayer} are satisfied for $X^{+}$ and $X^{-}$, then those must be satisfied for the sets $C^{+}$ and $C^{-}$. For convenience, let $k_X$ be $k$ in \eqref{eq:k} for $X^+$ and $X^-$ and let $k_C$ be $k$ in \eqref{eq:k} where we replace $X^+$ and $X^-$ by $C^+$ and $C^-$. We check the following:
\begin{itemize}
    \item The dimension $d$ doesn't change since $C^+, C^-\subset \R^d$, so $d$ is preserved.
    \item By \cref{def:complexity}, $C^+$ and $C^-$ are $\delta$-separated finite sets, so $\delta$ is preserved.
    \item By \cref{def:complexity}, $X_i\neq\emptyset$ and $c_i\in \conv(X_i)$ for every $i$. Hence, \cref{lem:geo lm}(2) implies $\Vert c_i\Vert \leq \Vert x\Vert \leq R$ for some $x\in X_i$. Thus, $C^+, C^-\subset R\B_2^d$, which means $R$ is preserved.
\end{itemize}
Finally, we check $\gamma$ satisfies the input condition now on $C^+$ and $C^-$ if it satisfies that for $X^+$ and $X^-$. Since $c_i\in \conv(X_i)$ for every $i$, \cref{lem:geo lm}(2) implies for every vector $g$ and indices $i$ with $\sigma(c_i)=+1$ and $\sigma (c_j)=-1$, there exists $x_i\in X_i$ and $x_j\in X_j$ such that $g^{\intercal}x_i\geq g^{\intercal}c_i$ and $(-g)^{\intercal}x_j\geq (-g)^{\intercal}c_j$. Hence
\[     w(C^+)
 = \E  \sup\lbrace g^{\intercal}(c_i) :\sigma(c_i)=+1\rbrace 
\leq \E  \sup\lbrace g^{\intercal}(x_i)\rbrace 
\leq \E  \sup\lbrace g^{\intercal}x : x\in X^+ \rbrace 
= w(X^+)\]
and similarly $w(C^-)\leq w(X^-)$, where with expectation is taken over $g$. Now
\begin{align*}
    w(C^+-C^-) & = \E  \sup \lbrace g^{\intercal}c : {c\in C^+-C^-}\rbrace 
    \\ & = \E  \sup\lbrace g^{\intercal}(c_i-c_j) :\sigma(c_i)=+1, \sigma (c_j)=-1 \rbrace
    \\ & \leq \E  \sup\lbrace g^{\intercal}(x_i-x_j) : x_i \in X_i, x_j \in X_j\rbrace 
    \\ & \leq \E  \sup\lbrace g^{\intercal}x : x\in X^+-X^- \rbrace 
    \\ & = w(X^+-X^-).
\end{align*}
Since we showed $R, \delta, d$ are preserved, this implies $k_X\geq k_C$, we obtain that
\[\gamma \leq \max(\delta^2/8Rd, \delta^2/18R\sqrt{k_X})\leq \max(\delta^2/8Rd, \delta^2/18R\sqrt{k_C}).\]
Thus, as $\gamma$ satisfies its input condition for $X^+$ and $X^-$, it does for $C^+$ and $C^-$.

\paragraph{Proof of Part (ii)}
Here we show that the output of \cref{alg:det_onelayer} on $C^{+}$ and $C^{-}$ linearly separates the points from $X^{+}$ and $X^{-1}$ with a margin $\sqrt{N}\mathcal{M}(\gamma/2, N)$. By Part (i), we can run \cref{alg:det_onelayer} on $C^+$ and $C^-$ with parameter $\gamma$ to define $h_j$ for each $1\leq j\leq N$ as in line 6. Let * denote unimportant reals. By definition of $h_j$, we obtain the exact same statement as \eqref{eq:hj} but for $C^+$ and $C^-$, i.e.
\begin{equation}\label{eq:hjci}
h_{j}(c_{i})=\begin{cases}
\leq -\gamma & \textup{if $j<i$ and $\sigma(c_j) \neq \sigma(c_i) $}\\
\geq \gamma  & \textup{if $j=i$}\\
* & \textup{otherwise}
\end{cases}.
\end{equation}
As $r_i\leq \gamma/2$ by \eqref{eq:ri}, for every $i, j$ and $x_{i}\in \B_2^d(c_{i}, r_{i})$, we have 
\[ 
    \vert h_{j}(x_{i})-h_{j}(c_{i})\vert =\vert w_{j}^{\intercal}(x_{i}-c_{i})\vert\leq \Vert x_{i}-c_{i}\Vert\leq r_{i}\leq \frac{\gamma}{2}.
\]
This gives an analog of \eqref{eq:hjci} on the entire sets $X_i$, i.e. for every $x_i\in X_i$
\begin{equation}\label{eq:hjxi}
h_{j}(x_{i})=\begin{cases}
\leq -\gamma/2 & \textup{if $j<i$ and $\sigma(x_j) \neq \sigma(x_i) $}\\
\geq \gamma/2  & \textup{if $j=i$}\\
* & \textup{otherwise}
\end{cases}.
\end{equation}
Hence, after activation, for every $x_i\in X_i$
\begin{equation}\label{eq:phi}
(\Phi(x_i))_j=\Re(h_{j}(x_{i}))=\begin{cases}
0  & \textup{if $j<i$ and $\sigma(x_j) \neq \sigma(x_i) $}\\
\geq \gamma/2  & \textup{if $j=i$}\\
\geq 0 & \textup{otherwise}
\end{cases}.
\end{equation}

We now proceed exactly as the proof of \cref{thm:arb}, except since $x_i$ can be any point in cluster $X_i$, we use supremum over $X_i$ instead of just $x_i$. Then, the proof will show the clusters $X_i$ are linearly separable according to $\sigma(x_i)$, which suffices since the clusters partition $X^+\cup X^-$. We choose the weight vector $a\in\R^N$ by picking $a_{i}$ recursively for $i=N, N-1, \dots , 1$ as follows:
\[ a_{i}=\frac{2\sigma(c_i) }{\gamma }\sup_{x\in X_i}\left(1+\left|\sum_{j=i+1}^{N}a_{j}(\Phi(x_{i}))_{j}\right|\right)\]
and so for hyperplane o$H(z)=a^{\intercal}z$, every $x_i\in X_i$, and every $i\in [N]$,
\begin{align*}
\sigma(x_i)  H(\Phi(x_{i})) & = \sigma(x_i) \sum_{j=1}^{N}a_{j}(\Phi(x_{i}))_{j}
\\ & = \sigma(x_i) \sum_{j<i, \sigma(x_j) =\sigma(x_i) }a_{j}(\Phi(x_{i}))_{j} + \sigma(x_i) \sum_{j<i, \sigma(x_j) \neq \sigma(x_i) }a_{j}(\Phi(x_{i}))_{j}
\\ & \qquad + \sigma(x_i) a_{i}(\Phi(x_{i}))_{i} + \sigma(x_i) \sum_{j>i}a_{j}(\Phi(x_{i}))_{j}
\end{align*}
by breaking into four cases similar to the proof of \cref{thm:fin}. The first term is non-negative since $a_j$ has the same sign as $\sigma(x_j)=\sigma(x_i)$ and coordinates of $\Phi(x_i)$ are non-negative due to $\Re$-activation. The second term is zero since $(\Phi(x_i))_j=0$ for such $j$ by \eqref{eq:phi}. Hence
\begin{align*}
\sigma(x_i)  H(\Phi(x_{i})) & \geq 0 + 0 + \frac{2\sigma(x_i) ^{2}}{\gamma}\sup_{x\in X_i}\left(1+\left|\sum_{j>i}a_{j}(\Phi(x_{i}))_{j}\right|\right)\left(\frac{\gamma}{2}\right)+\sigma(x_i) \sum_{j>i}a_{j}(\Phi(x_{i}))_{j}
\\ & = \sup_{x\in X_i}\left( 1+\left|\sum_{j>i}a_{j}(\Phi(x_{i}))_{j}\right|\right)+\sigma(x_i) \sum_{j>i}a_{j}(\Phi(x_{i}))_{j}
\\ & \geq 1
\end{align*}
by the triangle inequality. Thus, $H (\Phi(x_{i}))\geq 1$ if $x_{i}\in X^{+}$ and $H (\Phi(x_{i}))\leq -1$  if $x_{i}\in X^{-}$, so $H(z)=a^{\intercal}z$ separates $\Phi (X^{+})$ and $\Phi (X^{-})$. The margin of separation is at least $1/\Vert a\Vert$. We bound this as follows. Exactly as \eqref{eq:hiy}, $\vert h_{i}(y)\vert \leq 2R$ for every $y\in R\B_2^d$. Now, we bound each weight $a_{i}$
\[
\vert a_{i}\vert = \frac{2}{\gamma }\sup_{x\in X_i}\left( 1+\left|\sum_{j=i+1}^{N}a_{j}(\Phi(x_{i}))_{j}\right|\right)
\leq \frac{2}{\gamma} \left(1+2R \sum_{j=i+1}^{N}\vert a_{j}\vert\right).
\]

Thus, $\vert a_{i}\vert \leq \tilde{a}_{i}$ where $\tilde{a}_{i}$ is defined recursively by $\tilde{a}_{N}= 2/\gamma = \vert a_{N}\vert$ and 
\[
    \tilde{a}_{i} = \frac{2}{\gamma}\left(1 + 2R \sum_{j=i+1}^N \tilde{a}_j\right).
\]
We compute $\tilde{a}_{i}$ explicitly. Taking the successive differences
\[
    \tilde{a}_{i-1}-\tilde{a}_{i} = \frac{2}{\gamma}\left(1 + 2R \sum_{j=i}^N \tilde{a}_j\right) - \frac{2}{\gamma}\left(1 + 2R \sum_{j=i+1}^N \tilde{a}_j\right) = \left(\frac{4R}{\gamma}\right)\tilde{a}_{i},
\]
and using the fact that $\tilde{a}_{N}= 2/\gamma$, we get
\[
    \tilde{a}_{i} = \frac{2}{\gamma}\left(1+\frac{4R}{\gamma}\right)^{N-i}.
\]
Hence,
\[
\Vert a\Vert ^{2} \leq \sum_{i=1}^{N}\frac{4}{\gamma^{2}}\left(1+\frac{4R}{\gamma}\right)^{2(N-i)} = \frac{4}{\gamma^{2}} \cdot \frac{(1+4R/\gamma)^{2N}-1}{(1+4R/\gamma )^{2}-1} = \frac{(1+2R/(\gamma/2))^{2N}-1}{4R(R+\gamma/2)}.
\]
Rearranging gives desired bound on the margin: $1/\Vert a\Vert\ge \sqrt{N}\mathcal{M}(\gamma/2, N)$.
\end{proof}
\subsection{Passing to the Case of Random Initialization}
To generalize \cref{thm:fin} to one-layer RINN $\Phi$, we show that given sufficient width, for every $i$, with high probability there is a node in $\Phi$ that defines a hyperplane satisfying the conditions for $h_{i}$ in \cref{alg:det_onelayer}. Then, those nodes collected over all $i$ will make $X^{+}$ and $X^{-}$ separable, which will be enough for $\Phi$ to also make them separable. This amounts to the following two conditions:
\begin{enumerate}
\item The condition for each $h_{i}$ is satisfied by a random hyperplane with non-trivial probability, as shown in \cref{thm:prob}.
\item The other hyperplanes that do not fulfill the conditions for any $h_{i}$ do not hurt separation, as shown in \cref{lem:proj}.
\end{enumerate}

Combining these ideas, we are able to prove our main results (\cref{thm:main,thm:main in k}). Again, let us sketch the main ideas (of the finite $X^-$ and $X^+$ only), and hightlight a key obstacle, before the rigorous proofs in \cref{sec:1.3}.

For every $i\in [N]$ and $\ell\in [n]$, let $B_i^\ell$ be the event that the $\ell$-th random hyperplane $H_\ell$ of $\Phi$ satisfies the condition for $h_i$ on line 5 of \cref{alg:det_onelayer}.

We claim that $\Phi(X^+)$ and $\Phi(X^-)$ are separable if, for every $i$, $B_i^\ell$ holds for some $\ell$, i.e. $(\forall i)(\exists \ell)B_{i}^{\ell})$. If so, we can pick some hyperplanes in $\Phi$ to form an output of \cref{alg:det_onelayer} on $X^+, X^-$, and $\gamma$. We proceed as \cref{thm:fin} to show these hyperplanes make $X^+$ and $X^-$ separable via \cref{alg:weights}.

However, it is possible that the same $H_\ell$ may satisfy the conditions of multiple $h_i$'s, so we slightly modify \cref{alg:weights} to fix this issue. This modification shrinks the separation margin by a factor of $\sqrt{N}$ compared to \cref{thm:fin,thm:arb}. Then, by \cref{lem:proj}, $\Phi$ makes $X^+$ and $X^-$ separable with the desired margin, thus proving the claim. 

Using \cref{thm:prob}, we see that, for every $i$ and $\ell$, $\P (B_{i}^{\ell})\geq p$ or $p/10$ or $q$ depending on the distribution of $\Phi$. Applying indepedence over $\ell$ and union bound over $i$, we do a standard computation to bound $\P((\forall i)(\exists \ell)B_{i}^{\ell})$, thereby showing the finite set cases for \cref{thm:main,thm:main in k}.

For arbitrary sets $X^+$ and $X^-$, we modify apply the argument above to the centers $C^{+}$ and $C^{-}$ as guaranteed by \cref{def:complexity} to show $\Phi(C^{+})$ and $\Phi(C^{-})$ are separable. The result for $\Phi(X^{+})$ and $\Phi(X^{-})$ will then follow \cref{thm:arb}.
\subsection{Proof of \cref{thm:main,thm:main in k}}\label{sec:1.3}
We modify the notation slightly for convenience of proving both finite set cases and arbitrary set cases together. Fix $\gamma$ to be either $\delta^2/8Rd$ or $\delta^2/18R\sqrt{k}$ depending on the distribution of $\Phi$, as in the finite set cases. For the arbitrary set cases, the parameter will be $\gamma/2$. We show in all three cases that provided the respective width bounds are met, with probability $1-\eta$, $\Phi(X^+)$ and $\Phi(X^-)$ are linearly separable. If $\vert X^+\vert +\vert X^-\vert = N <\infty$, then $\Phi(X^+)$ and $\Phi(X^-)$ will be linearly separable with margin at least $\mathcal{M}(\gamma, N)$. Otherwise, in the case of arbitrary sets, $\Phi(X^+)$ and $\Phi(X^-)$ will be linearly separable with margin at least $\mathcal{M}(\gamma/2, N)$ where $N$ is now the $(\delta, 8R^2/\gamma)$-mutual complexity. We can check it is so for $\gamma\in\lbrace \delta^2/8Rd, \delta^2/18R\sqrt{k}\rbrace$.

For finite sets $X^+$ and $X^-$, recall $x_i$ and $\sigma$ from \cref{alg:det_onelayer} and \eqref{eq:sigma}. For arbitrary sets $X^+$ and $X^-$, let $C^{+}$ and $C^{-}$ be the finite sets of centers guaranteed by \cref{def:complexity} for  $(\delta, 8R^2/\gamma)$-mutual complexity. Let $c_{i}$ be the point in $C^{+}\cup C^{-}$ with the $i$-th largest norm and let $\sigma : C^{+}\cup C^{-}\to \lbrace -1, +1\rbrace$ such that $\sigma (c)=+1$ if $c\in C^{+}$ and $\sigma (c)=-1$ if $c\in C^{-}$.

For a random layer $\Phi(x)=\Re (Wx+b)$ with width $n$, we let $B_i^j$ for $1\leq i\leq N$ and $1\leq \ell\leq n$ be the event that the $\ell$-th random hyperplane $H_\ell(z)=W_\ell^{\intercal}z+b_\ell$ of $\Phi$ satisfies the condition for $h_i$ in \cref{alg:det_onelayer} applied to $X^+$ and $X^-$ when they are finite or on $C^+$ and $C^-$ otherwise, and parameter $\gamma$. More precisely, if $X^+$ and $X^-$ are finite sets, let
\begin{equation}\label{eq:bxil}
B^{\ell}_{i}(X):=\begin{cases}
\left\lbrace H_{\ell}(x_{i})\geq \gamma \right\rbrace\cap \bigcap_{j:j>i, \sigma(x_i) \neq\sigma(x_j)  }\left\lbrace H_{\ell}(x_{j})\leq -\gamma \right\rbrace & \textup{if such $j$ exists} \\
\left\lbrace H_{\ell}(x_{i})\geq \gamma \right\rbrace \cap \lbrace\exists \ x\in R\B_2^d: H_\ell(x)\leq 0\rbrace &\textup{otherwise}
\end{cases}.
\end{equation}
For arbitrary sets $X^+$ and $X^-$, analogously define
\begin{equation}\label{eq:bcil}
B^{\ell}_{i}(C):=\begin{cases}
\left\lbrace H_{\ell}(c_{i})\geq \gamma \right\rbrace\cap \bigcap_{j:j>i, \sigma(c_i) \neq\sigma(c_j)  }\left\lbrace H_{\ell}(c_{j})\leq -\gamma \right\rbrace & \textup{if such $j$ exists} \\
\left\lbrace H_{\ell}(c_{i})\geq \gamma \right\rbrace \cap \lbrace\exists \ c\in R\B_2^d: H_\ell(c)\leq 0\rbrace &\textup{otherwise}
\end{cases}
\end{equation}
with $\gamma$ chosen to be either $\delta^2/8Rd$ or $\delta^2/18R\sqrt{k}$ depending on the distribution of $\Phi$. This is similar to $B_i$ in \cref{thm:prob}, except we add the case where $j$ does not exist on line 5 of \cref{alg:det_onelayer}. We now resolve the duplication issue mentioned in the proof sketch.

\begin{claim}\label{claim:bilx}
Fix $\gamma$ to be $\delta^2/8Rd$ or $\delta^2/18R\sqrt{k}$. Recall $\mathcal{M}$ from \eqref{eq:p}. Then
\begin{enumerate}
    \item Suppose that $X^+$ and $X^-$ are finite and that for every $i$, $B_i^\ell(X)$ with parameter $\gamma$ holds for some $\ell$. Then, $\Phi(X^+)$ and $\Phi(X^-)$ are linearly separable with margin $\mathcal{M}(\gamma, N)$.
    \item For arbitrary sets $X^+$ and $X^-$, suppose that for every $i$, $B_i^\ell(C)$ with parameter $\gamma$ holds for some $\ell$. Then, $\Phi(X^+)$ and $\Phi(X^-)$ are linearly separable with margin $\mathcal{M}(\gamma/2, N)$.
\end{enumerate}

\end{claim}
\begin{proof}[Proof of \cref{claim:bilx}]
We break this proof down into four steps which we outline here. First, we set up some notation to clarify why we cannot directly apply \cref{alg:weights}. Then, we use \cref{thm:fin,thm:arb} to show that if we allow a single random hyperplane $H_\ell(z)$ to satisfy the condition for two hyperplanes $h_{i_1}$ and $h_{i_2}$, it makes the sets separable. Next, we show how to remove one of such hyperplanes while still making the sets separable. Finally, we use \cref{lem:proj} to show $\Phi$ also makes the sets separable. The margin will be tracked throughout.

\paragraph{Step 1: Setup}
By assumption, we can define indices map $f:[N]\to [n]$ by letting $f(i)=\ell$ if $B_i^\ell$ holds, where $B_i^\ell$ is either $B_i^\ell (X)$ or $B_i^\ell (C)$ depending on whether $X^+$ and $X^-$ are finite. For a given $i$, if there are multiple $\ell$ such that $B_i^\ell$ holds, define $f(i)$ to be any such $\ell$. Define the range $\Sigma :=f([N])\subset [n]$ and let $\pi:\R^n\to\R^N$ be the projection onto the coordinates indexed by $\Sigma$. When $f$ is not injective, we cannot apply \cref{thm:fin} directly to $\lbrace H_j:j\in\Sigma\rbrace$ since there are less than $N$ hyperplanes.  Let $h_i:=H_{f(i)}$ with another  \emph{duplicated} hyperplane $h_{i'}$, i.e. $h_i$ and $h_{i'}$ are two copies of $H_j$ if $f(i)=f(i')=j$. Then, $h_i$ satisfies the condition on line 5 of \cref{alg:det_onelayer} for this $i$. Consider the one-layer neural network $\Psi(x)=\Re(h_1(x), \dots, h_N(x))$.

\paragraph{Step 2: Showing $\Psi$ Separates}
To show $\Phi$ makes $X^+$ ad $X^-$ separable, we first need to show $\Psi$ does. For \cref{claim:bilx}(1), we first not that $\Psi$ is a valid output of \cref{alg:det_onelayer} on $X^+, X^-,$ and the chosen $\gamma$. Then, by \cref{thm:fin}, we can run \cref{alg:weights} to find vector $a\in \R^N$ such that the hyperplane $H(z):=a^{\intercal}z$ separates $\Psi(X^+)$ and $\Psi(X^-)$ with the margin at least $\sqrt{N}\mathcal{M}(\gamma, N)$, i.e.
\begin{equation}\label{eq:Hhxi}
H(\Psi(x))=\sum_{i=1}^N a_i \Re(h_i(x)) =\begin{cases} \geq \Vert a\Vert \sqrt{N}\mathcal{M}(\gamma, N) &\textup{if }x\in X^+ \\ \leq - \Vert a\Vert\sqrt{N} \mathcal{M}(\gamma, N) &\textup{if }x\in X^-
\end{cases}.
\end{equation}

For \cref{claim:bilx}(2), we first note that $\Psi$ is a valid output of \cref{alg:det_onelayer} on $C^+, C^-,$ and the chosen $\gamma$. By \cref{thm:arb}, $\Psi(X^+)$ and $\Psi(X^-)$ are linearly separable with margin at least $\sqrt{N}\mathcal{M}(\gamma/2, N)$. In fact, by the proof of \cref{thm:arb}, the separating hyperplane is of the form $H(z)=a^{\intercal}z$, i.e.
\begin{equation}\label{eq:Hhci}
H(\Psi(x))=\sum_{i=1}^N a_i \Re(h_i(x)) =\begin{cases} \geq \Vert a\Vert \sqrt{N}\mathcal{M}(\gamma/2, N) &\textup{if }x\in X^+ \\ \leq - \Vert a\Vert\sqrt{N} \mathcal{M}(\gamma/2, N) &\textup{if }x\in X^-
\end{cases}.
\end{equation}

\paragraph{Step 3: Showing $\pi\circ\Phi$ Separates}
Define real vector $\tilde{a}\in\R^\Sigma$ by
\[\tilde{a}_j := \sum_{i:f(i)=j}a_i \]
for all $j\in \Sigma$. In words, $\tilde{a}_j$ is the sum of the weights $a_i$ over all $i$ where $h_i$ is a duplicate of $H_j$. Now, we show that the hyperplane $\tilde{H}(z):=\tilde{a}^{\intercal}z$ separates $\pi(\Phi(X^+))$ and $\pi(\Phi(X^-))$ with margin at least $\mathcal{M}(\gamma, N)$. Note, for $j\in\Sigma$, $(\pi(\Phi(x)))_j = (\Phi(x))_j = \Re(H_j(x))$, and $H_j = h_i$ if $f(i)=j$. Hence
\[
    \tilde{H}(\pi(\Phi(x))) = \tilde{a}^{\intercal}\pi(\Phi(x)) = \sum_{j\in\Sigma} \tilde{a}_j(\pi(\Phi(x)))_{j} = \sum_{j\in\Sigma} \left(\sum_{i:f(i)=j}a_i\right)\Re(H_j(x))
\]
by the definition of $\tilde{a}_j$. Since $H_j=h_i$ if $f(i) = j$, we get
\[
\tilde{H}(\pi(\Phi(x))) = \sum_{j\in\Sigma} \sum_{i:f(i)=j}a_i\Re(h_i (x)) = \sum_{i=1}^Na_i\Re(h_i(x)) = H(\Psi(x))
\]
by \eqref{eq:Hhci}. By \eqref{eq:Hhxi} and \eqref{eq:Hhci}, $(\pi\circ \Phi)(X^+)$ and $(\pi\circ \Phi)(X^-)$ are linearly separable with margin at least $\Vert a\Vert \sqrt{N}\mathcal{M}(\gamma, N)/\Vert \tilde{a}\Vert$ if $X^+$ and $X^-$ are finite, and margin at least $\Vert a\Vert \sqrt{N}\mathcal{M}(\gamma/2, N)/\Vert \tilde{a}\Vert$ otherwise. For each $j$, by Cauchy-Schwarz on each vector $(a_i)_i$ indexed by $i\in f^{-1}(j)\subset [N]$
\[ \sum_{j\in\Sigma} \tilde{a}_j^2 = \sum_{j\in\Sigma} \left(\sum_{i:f(i)=j}a_i\right)^2 \leq \sum_{j\in\Sigma} \vert f^{-1}(j)\vert \sum_{i:f(i)=j}a_i^2 \leq N \sum_{j\in\Sigma} \sum_{i:f(i)=j}a_i^2= N \sum_{i=1}^N a_i^2, \]
where for a given $j$, $|f^{-1}(j)|$ is the number of indices $i$ such that $f(i) = j$. From this inequality, we get$\Vert \tilde{a}\Vert \leq \sqrt{N}\Vert a\Vert$. Thus, the separation margins are at least $\mathcal{M}(\gamma, N)$ and $\mathcal{M}(\gamma/2, N)$, respectively.

\paragraph{Step 4: Showing $\Phi$ Separates}
By \cref{lem:proj}, $\Phi(X^+)$ and $\Phi(X^-)$ are linearly separable with margin at least $\mathcal{M}(\gamma, N)$ or $\mathcal{M}(\gamma/2, N)$, depending on if $X^+$ and $X^-$ are finite. This proves \cref{claim:bilx}.
\end{proof} 

To show \cref{thm:main,thm:main in k}, it suffices to bound the probability that for every $i$, $B_i^\ell(X)$ and $B_i^\ell(C)$ occurs for some $\ell$, as in the following claim.

\begin{claim}\label{claim:bil prob}
For any $\delta$-separated sets $X^+, X^-\subset R\B_2^d$ with $N= \vert X^+\vert +\vert X^-\vert <\infty$, define $\Phi$ as in the three cases in \cref{thm:main,thm:main in k}, and define
$B_i^\ell(X)$ as in \eqref{eq:bxil}. Then, $\P\left((\forall i)(\exists \ell)B_{i}^{\ell}(X)\right)\geq 1-\eta$ provided the width bounds on $n$ from \cref{thm:main,thm:main in k} are met in each case.
\end{claim}
\begin{proof}[Proof of \cref{claim:bil prob}]
We break this proof down into two steps which we outline here. First, we will obtain bounds of $\P(B_i^\ell(X))$ similar to \cref{thm:prob}, except we include the case where no such $j$ exists as in \eqref{eq:bxil} and \eqref{eq:bcil}. Then, we use these bounds to obtain a final bound on $\P\left((\forall i)(\exists \ell)B_{i}^{\ell}(X)\right)$. 

\paragraph{Step 1: Bounding Each $\P(B_i^\ell(X))$}
If $j$ exists in \eqref{eq:bxil}, \cref{thm:prob} gives bounds $p$, $p/10$, and $q$ of $\P(B_i^\ell(X))$ for any $i$ and $\ell$, depending on $\Phi$ as in \cref{thm:main,thm:main in k}. We show the same result for the easier case where no such $j$ exists by reducing it to \cref{thm:prob}. If no such $j$ exists
\[ B^{\ell}_{i}(X) =
\left\lbrace H_{\ell}(x_{i})\geq \gamma \right\rbrace \cap \lbrace\exists \ x\in R\B_2^d: H_\ell(x)\leq 0\rbrace. \]
Since $\delta\leq 2R$, we can pick any $x_j\in R\B_2^d$ such that $\Vert x_j-x_i\Vert \geq \delta$. Then
\[\P\left(\left\lbrace H_{\ell}(x_{i}) \geq \gamma \right\rbrace \cap \lbrace\exists  \ x\in R\B_2^d: H_\ell(x)\leq 0\rbrace\right)
\geq \P\left(\left\lbrace H_{\ell}(x_{i})\geq \gamma \right\rbrace \cap \lbrace H_\ell (x_j)\leq -\gamma\rbrace\right). \] 
If $\Vert x_i\Vert \geq \Vert x_j\Vert$, then this is precisely event $B_i$ in \cref{thm:prob} where we want the random hyperplane $H_\ell$ to separate $x_i$ from the singleton set $\lbrace x_j\rbrace$ with parameter $\gamma$. If $\Vert x_i\Vert \leq \Vert x_j\Vert$, then this is precisely event $B_j$ in \cref{thm:prob} where we want the random hyperplane $-H_\ell$ to separate $x_j$ from the singleton set $\lbrace x_i\rbrace$ with parameter $\gamma$. Then, by \cref{thm:prob}, $\P (B_{i}^{\ell}(X))\geq p$ or $p/10$ or $q$ for every $i$ and $\ell$, depending on the distribution of weights as in \cref{thm:main,thm:main in k}. 

\paragraph{Step 2: Bounding $\P\left((\forall i)(\exists \ell)B_{i}^{\ell}(X)\right)$}
We show \cref{claim:bil prob} when the distribution of $\Phi$ is from \cref{thm:main}(i). The cases for \cref{thm:main}(ii) and \cref{thm:main in k} follow by replacing $p$ with $p/10$ and $q$, respectively. For convenience, let $\overline{B}$ be the complement of $B$. By a union bound over $i$ and independence of $\ell\in [n]$
\begin{align*}
\P\left((\forall i)(\exists \ell)B_{i}^{\ell}(X)\right) & = 1-\P\left((\exists i)(\forall \ell)\overline{B_{i}^{\ell}(X)}\right)
\\ & \ge 1-N\P\left((\forall \ell)\overline{B_{i}^{\ell}(X)}\right)
\\ & \ge 1-N(1-p)^{n}
\\ & \ge 1-\eta
\end{align*}
where we check $1-x\leq e^{-x}\ \forall \ x\in [0, 1]$, so the last inequality follows $(1-p)^n\leq \exp(-pn)\leq \eta/N$ by our bound on $n$. This concludes the proof of \cref{claim:bil prob}.
\end{proof}
Finally, \cref{claim:bilx} and \cref{claim:bil prob} together implies the finite set cases of \cref{thm:main,thm:main in k}. For the arbitrary set cases, apply \cref{claim:bil prob} to the $\delta$-separated finite sets $C^+, C^-\subset R\B_2^d$ to get that $\P\left((\forall i)(\exists \ell)B_{i}^{\ell}(C)\right)\geq 1-\eta$ provided the width bounds on $n$ in \cref{thm:main,thm:main in k} are met. Then, \cref{claim:bilx} implies the arbitrary set cases of \cref{thm:main,thm:main in k}.
\section{Conclusion}
We showed that a sufficiently wide one-layer RINN transforms two $\delta$-separated sets in $\mathbb{R}^d$ into two linearly separable sets with high probability. We derived explicit bounds on the width of the network for this to happen in two distinct scenarios. One of our bounds depends exponentially on $d$, but polynomially on all the parameters of the weight distribution of the RINN, thereby improving on the existing bound. In comparison, our second bound depends polynomially on $d$, overcoming the curse of dimensionality. 


\section*{Acknowledgments}
YS was supported through the Simons Investigator Award grant (622132) of Elchanan Mossel and the UROP program of MIT. PG, SM and YS thank the productive research environment at the MIT mathematics department.
 



\bibliographystyle{imsart-number}		
\bibliography{ref}

\end{document}